\pdfoutput=1

\documentclass[11pt]{article}

\usepackage[preprint]{acl}

\usepackage{times}
\usepackage{latexsym}

\usepackage[T1]{fontenc}

\usepackage[utf8]{inputenc}

\usepackage{microtype}

\usepackage{inconsolata}

\usepackage{graphicx}

\usepackage{kotex}
\usepackage{color}
\usepackage{makecell}
\usepackage{xurl}
\usepackage{multirow}
\usepackage{subcaption}
\usepackage{booktabs}
\usepackage{xcolor, soul}
\usepackage{amssymb}
\usepackage{bbm}
\usepackage{tcolorbox}
\usepackage{amsmath}
\usepackage{tabularx}
\usepackage{longtable}

\definecolor{lightcyan}{rgb}{0.88,1,1}
\definecolor{lightyellow}{rgb}{1,1,0.7}
\definecolor{lightpurple}{rgb}{0.9,0.8,1}
\definecolor{lightgreen}{rgb}{0.8,1,0.8}
\newcommand{\hlcyan}[1]{\sethlcolor{lightcyan}\hl{#1}}
\newcommand{\hlyellow}[1]{\sethlcolor{lightyellow}\hl{#1}}
\newcommand{\hlpurple}[1]{\sethlcolor{lightpurple}\hl{#1}}
\newcommand{\hlgreen}[1]{\sethlcolor{lightgreen}\hl{#1}}

\title{Hypothetical Documents or Knowledge Leakage?\\ Rethinking LLM-based Query Expansion}

\author{
Yejun Yoon$^{\heartsuit}$~~~~~~Jaeyoon Jung$^{\clubsuit\diamondsuit}$~~~~~~Seunghyun Yoon$^{\spadesuit}$~~~~~~Kunwoo Park$^{\clubsuit\heartsuit}$\\ 
$^{\heartsuit}$Department of Intelligent Semiconductors, Soongsil University\\
$^{\clubsuit}$School of AI Convergence, Soongsil University\\
$^{\diamondsuit}$MAUM AI Inc.\\
$^{\spadesuit}$Adobe Research, USA\\
\texttt{\{yejun0382, jaeyoonskr\}@soongsil.ac.kr}, \texttt{syoon@adobe.com}, \texttt{kunwoo.park@ssu.ac.kr}
}

\begin{document}
\maketitle
\begin{abstract}
Query expansion methods powered by large language models (LLMs) have demonstrated effectiveness in zero-shot retrieval tasks. These methods assume that LLMs can generate hypothetical documents that, when incorporated into a query vector, enhance the retrieval of real evidence. However, we challenge this assumption by investigating whether knowledge leakage in benchmarks contributes to the observed performance gains. Using fact verification as a testbed, we analyze whether the generated documents contain information entailed by ground-truth evidence and assess their impact on performance. Our findings indicate that, on average, performance improvements consistently occurred for claims whose generated documents included sentences entailed by gold evidence. This suggests that knowledge leakage may be present in fact-verification benchmarks, potentially inflating the perceived performance of LLM-based query expansion methods.
\end{abstract}

\section{Introduction}

Zero-shot retrieval aims to identify relevant documents without requiring any relevance supervision for training a retriever~\cite{10.1145/3637870}. Because obtaining query-document pairs, such as MS-MARCO~\cite{bajaj2016ms}, for supervised training is challenging, developing zero-shot retrieval methods is both difficult and highly desirable for effectively addressing knowledge-intensive applications~\cite{lewis2020retrieval}, including question answering~\cite{zhu2021retrieving} and fact verification~\cite{fact_checking_survey}. 

Recent studies have leveraged the natural language generation capabilities of large language models (LLMs) to enhance the performance of zero-shot retrieval~\cite{beir}. LLM-based query expansion (QE) uses LLMs to generate documents that extend a query~\cite{jagerman2023query,lei2024corpus,10.1145/3539618.3591992}. Approaches such as HyDE~\cite{gao-etal-2023-precise} and Query2doc~\cite{wang-etal-2023-query2doc}, which have been widely adopted in recent research~\cite{wang-etal-2024-searching, chen-etal-2024-analyze, yoon-etal-2024-hero}, have achieved notable performance gains across various benchmarks without retriever parameter updates. These approaches prompt LLMs to generate documents that answer a question or verify a claim. Although these generated documents, referred to as \emph{hypothetical} documents, may contain factual errors or hallucinations, it is assumed that incorporating them into a query can enhance retrieval of relevant \emph{real} documents~\cite{gao-etal-2023-precise}. 

\begin{figure}[t]
\includegraphics[width=0.99\linewidth]{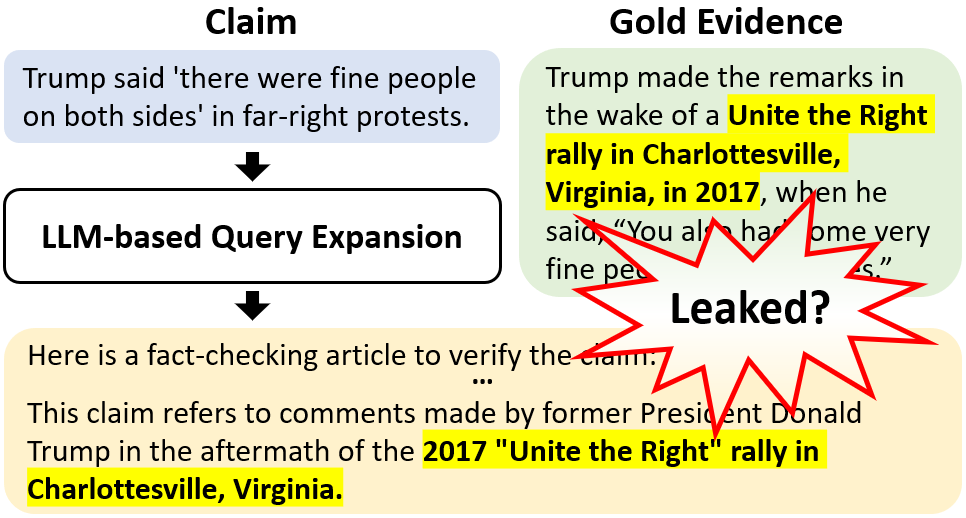}
\caption{Illustration of potential knowledge leakage in LLM-based query expansion.}
\label{fig:query_expansion}
\end{figure}

In this paper, we challenge the underlying assumption, as illustrated by a counterexample in Figure~\ref{fig:query_expansion}: \emph{Do LLMs truly generate hypothetical documents, or are they merely reproducing what they already know?} LLMs are extensively pretrained on vast corpora, primarily collected from the web. As common retrieval targets, such as Wikipedia and web documents, are often included in these pretraining corpora~\cite{groeneveld-etal-2024-olmo, touvron2023llama, du2022glam, brown2020language}, many available LLMs may already contain knowledge relevant to a given query and retrieval targets in benchmark datasets, a phenomenon we refer to as \emph{knowledge leakage}. If knowledge leakage occurs and an LLM reproduces what it already knows about a given query, its generation cannot be seen as hypothetical documents. Furthermore, assessing QE performance on a benchmark where leakage occurs could lead to overestimation, as retrieving verbatim or semantically equivalent sentences is a much more trivial task than predicting the relevance between a query and document when the relevance signal is not explicit.

To investigate whether knowledge leakage exists in popular benchmarks and how it affects downstream performance, this study uses fact verification as a testbed. We analyze whether LLM-generated documents contain sentences entailed by gold evidence and assess their impact on performance. Across experiments spanning three benchmarks and seven LLMs, we observed a consistent trend: QE methods were effective, on average, only when the generated documents included sentences entailed by gold evidence. These findings suggest that knowledge leakage may be present in these benchmarks, potentially inflating the perceived effectiveness of LLM-based QE methods.

\section{Related Works}

\paragraph{LLM-based Query Expansion} QE has been explored as a means to enhance retrieval performance by enriching the initial query representation~\cite{query_expansion_survey}. One widely studied approach is relevance feedback~\cite{rocchio1971relevance, 10.1145/383952.383972,amati2002probabilistic}, which leverages feedback signals to expand the query. Recent work has explored LLM's generative capabilities for QE~\cite{zhu2023large,jagerman2023query,lei2024corpus}. \citet{mackie2023generative}, for instance, introduced a method that utilizes LLM-generated documents as relevance feedback. Meanwhile, other researchers proposed HyDE~\cite{gao-etal-2023-precise} and Query2doc~\cite{wang-etal-2023-query2doc}, which employs LLMs to generate hypothetical documents based on an initial query.
Despite their simplicity, these methods have demonstrated substantial effectiveness across benchmarks for zero-shot retrieval and knowledge-intensive tasks~\cite{wang-etal-2024-searching}, including fact verification~\cite{yoon-etal-2024-hero}.

\paragraph{Data Leakage and LLM Memorization} Previous research has investigated various forms of data leakage in LLMs~\cite{kandpal2023large, samuel-etal-2025-towards, deng-etal-2024-investigating, xu2024benchmark}. One study used perplexity to detect potential data leakage, uncovering substantial instances of training or even test set misuse~\cite{xu2024benchmarking}. \citet{deng2023benchmark} further examined data contamination by predicting masked tokens in test sets and found that GPT 3.5 could reconstruct missing portions of MMLU~\cite{hendrycks2020measuring} test instances with 57\% accuracy. Other research has explored the boundaries of LLM knowledge~\cite{yin-etal-2023-large, dong2024statistical, burns2022discovering,kadavath2022language,mallen-etal-2023-trust}, including a refusal-aware instruction tuning method that trains LLMs to reject uncertain questions~\cite{zhang-etal-2024-r}---where an LLM is deemed uncertain if its generated response does not match the ground truth. Another study leveraged response consistency to estimate an LLM's confidence in its knowledge~\cite{cheng2024can}. In this work, we apply NLI~\cite{maccartney2009natural} to LLM-generated documents, referencing gold evidence, to examine what an LLM knows.

\section{Methodology}

\subsection{Task and Dataset}
Fact verification aims to predict the veracity label of a textual claim $c$. Depending on the dataset, the veracity label can fall into one of three or four categories\footnote{AVeriTeC provides four categories, whereas FEVER and SciFact use three, excluding \textit{conflicting evidence}. In SciFact, \textit{CONTRADICT} is treated equivalently to \textit{refuted}.}: \textit{supported}, \textit{refuted}, \textit{not enough evidence}, or \textit{conflicting evidence}.

The fact-verification task consists of two sub-tasks: evidence retrieval and verdict prediction. In evidence retrieval, a retrieval method $R(\cdot)$ identifies an evidence set $\tilde{E}=\{\tilde{e_1},\cdots,\tilde{e_k}\}$ from a knowledge store $K$ (e.g., Wikipedia), used to verify $c$. The performance of $R$ is evaluated by comparing $\tilde{E}$ with $E=\{e_1,\cdots,e_{l}\}$, the gold evidence set. Verdict prediction then determines the veracity label of $c$ based on $\tilde{E}$. 

We choose fact verification as the target task to test our hypothesis for two main reasons. First, in real-world fact-checking scenarios~\cite{miranda2019automated,nakov2021automated}, retrieving evidence about niche or novel knowledge is crucial. If QE is effective only when the relevant knowledge has been seen during language model pretraining, its practical usefulness may be limited. Second, verdict prediction is a classification task, which enables clearer evaluation of how QE influences end performance compared to generation-based tasks, such as factual QA~\cite{joshi-etal-2017-triviaqa,10.1162/tacl_a_00276}. 

We employ three datasets that provide annotated evidence and corresponding veracity labels, along with an external knowledge store: \textsf{FEVER}~\cite{fever}, \textsf{SciFact}~\cite{scifact}, and \textsf{AVeriTeC}~\cite{schlichtkrull2023averitec}. While \textsf{FEVER} and \textsf{SciFact} contain verbatim extractions from $K$ as gold evidence (i.e., $E$), \textsf{AVeriTeC} uses human-written evidence referencing $K$ to verify claims. These three datasets cover diverse knowledge domains, as summarized in Table~\ref{tab:data}.

\subsection{LLM-based Query Expansion}

We consider two LLM-based QE methods.

\paragraph{Query2doc \cite{wang-etal-2023-query2doc}} generates a pseudo-document $d$ based on a query $q$. It then forms an expanded query $q^{+}$ by concatenating $d$ with multiple copies of $q$ (Equation \ref{eq:q2d}).
\begin{equation}
\label{eq:q2d}
    q^{+} = concat(q \times n, d)
\end{equation}
The expanded query $q^{+}$ is then used to retrieve documents via BM25~\cite{Lin_etal_SIGIR2021_Pyserini}. Following \citet{jagerman2023query}, we set $n$ as 5.

\paragraph{HyDE \cite{gao-etal-2023-precise}} employs an LLM to generate hypothetical documents $[d_1,...,d_N]$ in response to a query $q$. A dense retriever $g(\cdot)$ encodes $q$ and each $d_k$ separately, and their encoded embeddings are averaged to form the query vector $v_{q^+}$ (Equation \ref{eq:hyde}). 
\begin{equation}
\label{eq:hyde}
    v_{q^+} = \frac{1}{N+1} \sum_{k=1}^{N}[g(d_{k})+g(q)]
\end{equation}
Here, we set $N$ to 1 in this study. We use Contriever~\cite{contriever} as $g(\cdot)$ with prompts provided in Appendix~\ref{app:prompt}.

\subsection{Matching Algorithm}

We present an algorithm to determine whether an LLM-generated document $d$ for a claim $c$ contains a sentence that is verbatim or semantically equivalent to the gold evidence $E=\{e_1,\cdots,e_m\}$. If such a sentence exists, it may indicate that the backbone LLM was exposed to knowledge related to the corresponding evidence during training. To automatically identify such claims, we rely on natural language inference (NLI) assigning each claim $c$ to one of two conditions: \textit{matched} ($M$) or \textit{unmatched} ($\neg M$). The algorithm consists of three steps. 

\begin{itemize}
    \item[\textbf{(1)}]\textbf{Sentence Segmentation}: Segment $d$ into sentences and remove reproductions of $c$ to construct $S = \{s_1,\cdots,s_n\}$.
    \item[\textbf{(2)}]\textbf{NLI Labeling}: Annotate a label $l_{(i,j)}$ for each pair $(e_i,s_j) \in E \times S$, where $l_{(i,j)}\in\{$\textit{entailment}, \textit{contradiction}, \textit{neutral}$\}$.
    \item[\textbf{(3)}]\textbf{Label Aggregation}: Aggregate all labels $\{l_{(1,1) },\cdots,l_{(i,j)},\cdots,l_{(m,n)}\}$ into a single label $l$. If there exists $l_{(i,j)}$ labeled as \textit{entailment}, assign \textit{matched}, otherwise assign \textit{unmatched}.
\end{itemize}

We use the sentence segmentation module provided by spaCy\footnote{\url{https://huggingface.co/spacy/en_core_web_lg}}, and employ GPT-4o-mini for NLI (Figure~\ref{fig:app:NLI_prompt}). To filter out sentences of reproduced claims in LLM responses, we apply ROUGE-2~\cite{lin2004rouge} with a threshold of 0.95, based on manual inspection.

\begin{table}[t]
    \centering
    \small
    \resizebox{.99\linewidth}{!}{
    \begin{tabular}{cccc}
        \hline
        Dataset & \# claim & \makecell{\# gold evidence\\per claim} & \makecell{\# documents\\(knowledge source)} \\\hline
        FEVER & 6,666 & 1.66 & \makecell{5,416,536\\(Wikipedia)}\\
        SciFact & 693 & 1.8 & \makecell{5,183\\(Paper abstracts)}\\
        AVeriTeC & 3,563 & 2.77 &\makecell{2,623,538 \\(Web documents)}\\\hline
    \end{tabular}
    }
    \caption{Dataset statistics}
    \label{tab:data}
\end{table}

\section{Experimental Results}
\label{sec:results}

\begin{table*}[t]
\centering
\begin{subtable}{0.95\textwidth}
    \centering
    \resizebox{\textwidth}{!}{
    \begin{tabular}{c|ccc|ccc|ccc}
    \hline
    \multirow{2}{*}{Method} & \multicolumn{3}{c|}{FEVER} & \multicolumn{3}{c|}{SciFact} & \multicolumn{3}{c}{AVeriTeC} \\ \cline{2-10}
    & Recall@5 & NDCG@5 & F1 & Recall@5 & NDCG@5 & F1 & METEOR & BERTScore & F1\\ \hline
    BM25        & 31   & 25   & 54   & 51.2 & 45.5 & 48.6 & 17.8 & 11.6 & 32 \\\hline 
    \multicolumn{10}{c}{Performance by varying LLMs}\\\hline
    GPT-4o-mini      & 36.4$\pm$0.1  & 29.3$\pm$0.1  & 55.6$\pm$0.1  & 55.1$\pm$0.2  & 47.9$\pm$0.1  & \textbf{52.5$\pm$0.5}  & 19.1$\pm$0.0  & 12.4$\pm$0.0  & 32.6$\pm$0.1 \\
    Claude-3-haiku   & 35.2$\pm$0.1  & 28.3$\pm$0.1  & 55.4$\pm$0.1  & 56$\pm$0.1  & 48.2$\pm$0.1  & \textbf{52$\pm$0.5}  & 19.3$\pm$0.0  & 12.5$\pm$0.0  & 33.1$\pm$0.1 \\
    Gemini-1.5-flash & 36.2$\pm$0.1  & 29.2$\pm$0.1  & 55.8$\pm$0.1  & \textbf{56.2$\pm$0.2}  & \textbf{49.4$\pm$0.1}  & \textbf{52.2$\pm$0.5}  & 18.9$\pm$0.0  & 12.5$\pm$0.0  & \textbf{33.3$\pm$0.2}  \\\hline
    Llama-3.1-8b-it  & 35.7$\pm$0.1  & 28.6$\pm$0.2  & 55.6$\pm$0.2  & 54.9$\pm$0.2  & 47.8$\pm$0.2  & 51.9$\pm$0.3  & 19$\pm$0.0  & 12.4$\pm$0.0  & 32.2$\pm$0.2   \\
    Llama-3.1-70b-it & \textbf{38.3$\pm$0.1}  & \textbf{31$\pm$0.1}  & \textbf{56.1$\pm$0.1}  & \textbf{56.4$\pm$0.3}  & 49.2$\pm$0.1  & \textbf{52.4$\pm$0.7}  & \textbf{19.3$\pm$0.1}  & \textbf{12.7$\pm$0.0}  & \textbf{33.4$\pm$0.2}   \\
    Mistral-7b-it    & 35.1$\pm$0.3  & 28$\pm$0.2  & 55.4$\pm$0.2  & 55.1$\pm$0.1  & 47.9$\pm$0.1  & \textbf{51.9$\pm$0.6}  & 19.2$\pm$0.0  & 12.6$\pm$0.0  & 32.8$\pm$0.1   \\
    Mixtral-8x7b-it  & 35.1$\pm$0.2  & 27.9$\pm$0.2  & 55.3$\pm$0.2  & 54.6$\pm$0.2  & 47.7$\pm$0.1  & 51.9$\pm$0.4  & \textbf{19.4$\pm$0.0}  & \textbf{12.7$\pm$0.0}  & 33.2$\pm$0.1   \\ \hline
    \end{tabular}
    }
    \caption{Query2doc}
    \vspace{2mm}
\end{subtable}
\hfill
\begin{subtable}{0.95\textwidth}
    \centering
    \resizebox{\textwidth}{!}{
    \begin{tabular}{c|ccc|ccc|ccc}
    \hline
    \multirow{2}{*}{Method} & \multicolumn{3}{c|}{FEVER} & \multicolumn{3}{c|}{SciFact} & \multicolumn{3}{c}{AVeriTeC} \\ \cline{2-10}
    & Recall@5 & NDCG@5 & F1 & Recall@5 & NDCG@5 & F1 & METEOR & BERTScore & F1\\ \hline
    Contriever        & 26.8 & 20.2 & 53.1 & 55.1 & 47.3 & 51.2 & 17.6 & 12.6 & 33.9 \\\hline 
    \multicolumn{10}{c}{Performance by varying LLMs}\\\hline
    GPT-4o-mini      & 37.3$\pm$0.1  & 28.8$\pm$0.0  & 55.6$\pm$0.1  & 61.2$\pm$0.2  & 53.1$\pm$0.1  & \textbf{54$\pm$0.5}  & 18.7$\pm$0.0  & 13.2$\pm$0.0  & \textbf{35.7$\pm$0.6} \\
    Claude-3-haiku   & 36.7$\pm$0.1  & 28.1$\pm$0.0  & 55.6$\pm$0.1  & \textbf{62.8$\pm$0.1}  & \textbf{54.7$\pm$0.1}  & \textbf{53.7$\pm$0.4}  & \textbf{19.3$\pm$0.0}  & \textbf{14$\pm$0.0}  & \textbf{36.2$\pm$0.6} \\
    Gemini-1.5-flash & 35$\pm$0.1  & 26.7$\pm$0.1  & 55.2$\pm$0.2  & 61$\pm$0.2  & 52.9$\pm$0.2  & \textbf{53.5$\pm$0.7}  & 18$\pm$0.0  & 12.4$\pm$0.0  & \textbf{35.7$\pm$0.5}  \\\hline
    Llama-3.1-8b-it  & 36.7$\pm$0.1  & 28.4$\pm$0.1  & 55.4$\pm$0.1  & 61.2$\pm$0.2  & 53.4$\pm$0.2  & \textbf{53.6$\pm$0.7}  & 18.9$\pm$0.0  & 13.6$\pm$0.0  & 35.5$\pm$0.4   \\
    Llama-3.1-70b-it & \textbf{40.4$\pm$0.2}  & \textbf{31.7$\pm$0.2}  & \textbf{55.9$\pm$0.1}  & 61.9$\pm$0.3  & 54.1$\pm$0.2  & \textbf{53.6$\pm$0.5}  & 19$\pm$0.2  & 13.7$\pm$0.1  & 35.4$\pm$0.7   \\
    Mistral-7b-it    & 36.3$\pm$0.1  & 27.8$\pm$0.0  & 55.3$\pm$0.1  & 60.7$\pm$0.2  & 52.7$\pm$0.2  & 53.4$\pm$0.4  & 19$\pm$0.0  & 13.6$\pm$0.0  & \textbf{35.8$\pm$0.7}   \\
    Mixtral-8x7b-it  & 37.6$\pm$0.1  & 29.1$\pm$0.1  & 55.7$\pm$0.1  & 61.3$\pm$0.2  & 53.1$\pm$0.1  & 53.3$\pm$0.3  & 19.2$\pm$0.0  & 13.7$\pm$0.0  & \textbf{35.8$\pm$0.7}   \\ \hline
    \end{tabular}
    }
    \caption{HyDE}
\end{subtable}
\caption{Fact verification performance using baseline retrievers and LLM-based query expansion methods, with the number of retrieved evidence set to five ($k=5$). }
\label{tab:fact_verification_top_5}
\end{table*}

\begin{table*}[ht]
\centering
\resizebox{0.8\textwidth}{!}{
\begin{tabular}{c|cc|cc|cc}
\hline
\multirow{2}{*}{Method} & \multicolumn{2}{c|}{FEVER} & \multicolumn{2}{c|}{SciFact} & \multicolumn{2}{c}{AVeriTeC} \\ \cline{2-7}
& Query2doc & HyDE & Query2doc & HyDE & Query2doc & HyDE \\ \hline
GPT-4o-mini      & 75.8$\pm$0.1 & \textbf{83.5$\pm$0.1} & 52.8$\pm$0.4 & 59.1$\pm$0.7 & 40.4$\pm$0.2 & 68$\pm$0.3 \\
Claude-3-haiku   & 76.6$\pm$0.1 & 77.8$\pm$0.1 & 56.1$\pm$0.4 & 53.8$\pm$0.2 & 40.8$\pm$0.1 & 62.3$\pm$0.2 \\
Gemini-1.5-flash & 69.9$\pm$0.3 & 70.2$\pm$0.3 & 50.8$\pm$0.6 & \textbf{27.6$\pm$0.7} & 44.1$\pm$0.1 & 59.6$\pm$0.3 \\\hline
Llama-3.1-8b-it  & 68.5$\pm$1.0 & 73$\pm$0.9 & 53.9$\pm$0.5 & 48.2$\pm$0.6 & 37.4$\pm$1.1 & 53.8$\pm$1.0 \\
Llama-3.1-70b-it & 78.3$\pm$0.7 & 71.7$\pm$0.7 & 57.5$\pm$0.3 & 55$\pm$0.8 & 48.1$\pm$0.8 & 47$\pm$4.8 \\
Mistral-7b-it    & 72.5$\pm$0.2 & 75$\pm$0.2 & 51.1$\pm$0.5 & 49.4$\pm$0.8 & 44.7$\pm$0.2 & 55.6$\pm$0.3 \\
Mixtral-8x7b-it  & 78.7$\pm$0.1 & 81$\pm$0.1 & 55.9$\pm$0.6 & 54.7$\pm$0.7 & 49.4$\pm$0.3 & 56.7$\pm$1.3\\ \hline
\end{tabular}
}
\caption{Proportion of expanded queries containing sentences entailed by ground truth evidence.}
\label{tab:matched}
\end{table*}

We conducted evaluation experiments on three fact-verification benchmarks. Based on eight repetitions of LLM generation, we report the average performance along with the standard error.

\begin{table*}[t]
\centering
\resizebox{.95\textwidth}{!}{
\begin{tabular}{cc|ccc|ccc|ccc}
\hline
\multirow{2}{*}{Method} &\multirow{2}{*}{Data} & \multicolumn{3}{c|}{FEVER} & \multicolumn{3}{c|}{SciFact} & \multicolumn{3}{c}{AVeriTeC} \\ \cline{3-11}
& & Recall@5 & NDCG@5 & F1 & Recall@5 & NDCG@5 & F1 & METEOR & BERTScore & F1\\ \hline
BM25 & ALL & 31 & 25 & 54 & 51.2 & 45.5 & 48.6 & 17.8 & 11.6 & 32 \\\hline
\multirow{3}{*}{Query2doc} & ALL   & 36.4$\pm$0.1 & 29.3$\pm$0.1 & 55.6$\pm$0.1 & 55.1$\pm$0.2 & 47.9$\pm$0.1 & 52.5$\pm$0.5 & 19.1$\pm$0.0 & 12.4$\pm$0.0 & 32.6$\pm$0.1 \\
& $M$       & \textbf{40.5$\pm$0.1} & \textbf{32.8$\pm$0.1}  & \textbf{58.4$\pm$0.0} & \textbf{63.3$\pm$0.4} & \textbf{57.1$\pm$0.3}  & \textbf{53.7$\pm$0.3} & \textbf{21.6$\pm$0.1} & \textbf{17.6$\pm$0.1} & \textbf{38.3$\pm$0.3}  \\
& $\neg M$  & 23.8$\pm$0.3 & 18.5$\pm$0.2 & 44.9$\pm$0.1 & 45.9$\pm$0.4 & 37.6$\pm$0.3  & 49$\pm$0.4 & 17.4$\pm$0.0  & 9$\pm$0.0 & 27.6$\pm$0.1 \\ \hline
Contriever & ALL & 26.8 & 20.2 & 53.1 & 55.1 & 47.3 & 51.2 & 17.6 & 12.6 & 33.9 \\\hline 
\multirow{3}{*}{HyDE} & ALL  & 37.3$\pm$0.1 & 28.8$\pm$0.0 & 55.6$\pm$0.1 & 61.2$\pm$0.2   & 53.1$\pm$0.1 & 54$\pm$0.5 & 18.7$\pm$0.0 & 13.2$\pm$0.0 & 35.7$\pm$0.6 \\
 & $M$ & \textbf{40$\pm$0.1} & \textbf{30.9$\pm$0.1} & \textbf{57.8$\pm$0.0} & \textbf{68.4$\pm$0.3} & \textbf{61.4$\pm$0.3} & \textbf{57.1$\pm$0.2} & \textbf{19.8$\pm$0.0} & \textbf{15.5$\pm$0.0} & \textbf{37$\pm$0.2} \\
 & $\neg M$       & 23.4$\pm$0.4 & 17.9$\pm$0.3 & 41.9$\pm$0.2 & 50.8$\pm$0.5 & 41.2$\pm$0.4 & 48.9$\pm$0.4 & 16.4$\pm$0.0 & 8.3$\pm$0.1 & 30.3$\pm$0.4   \\ \hline
\end{tabular}}
\caption{Fact verification performance based on whether documents generated by query expansion methods with GPT-4o-mini contain sentences entailed by gold evidence. Results for other LLMs are presented in Table~\ref{tab:app:nli_top_5}.}
\label{tab:nli}
\end{table*}

\paragraph{Are LLM-based query expansion methods effective for fact verification?}

To assess the effectiveness of LLM-based QE methods, we compared their performance against BM25 and Contriever that use $c$ as query, respectively, as baseline retrievers. For evidence retrieval, we used Recall@$k$ and NDCG@$k$ ($k=5$) as evaluation metrics~\cite{manning2009introduction} on the FEVER and SciFact datasets, where both the ground-truth evidence $E$ and retrieved evidence $\tilde{E}$ come from the knowledge store $K$. In contrast, $\tilde{E}$ in AveriTeC consists of human-written evidence rather than extracts from $K$. Therefore, following previous studies~\cite{schlichtkrull2023averitec,chen-etal-2022-generating}, we applied the Hungarian algorithm~\cite{kuhn1955hungarian} with METEOR~\cite{banerjee-lavie-2005-meteor} and BERTScore~\cite{bert-score} on the top five retrieved sentences, computing token-level and embedding-level similarity, respectively, based on a binary assignment between generated and reference sequences. For verdict prediction, we used GPT-4o-mini with the five retrieved evidence and evaluated performance using macro F1. Evaluation details are provided in Appendix~\ref{sec:app:metrics}. 

As shown in Table~\ref{tab:fact_verification_top_5}, both Query2doc and HyDE consistently outperformed BM25 and Contriever across all three datasets and for seven different backbone LLMs (three proprietary and four open models). The performance gap between each baseline and its respective expansion method was statistically significant (\textit{p}$<$0.001), demonstrating the effectiveness of the QE methods for evidence retrieval and, consequently, verdict prediction in these benchmarks. Results for $k=10$ are available in Appendix~\ref{sec:app:results}, showing a consistent trend.

\paragraph{Do LLM-generated documents include ground truth evidence?}

Table~\ref{tab:matched} presents the proportion of matched claims observed for the three datasets by using seven different LLMs. In most cases, more than 40\% of the claims were matched, with a few exceptions. The lowest proportion (27.6\%) was observed for SciFact when claims were expanded using HyDE with Gemini-1.5-flash---still a notable fraction. The highest proportion (83.5\%) was observed for FEVER when using HyDE with GPT-4o-mini. Several examples of LLM-generated documents and gold evidence are provided in Table~\ref{tab:app:examples}.

\paragraph{How does performance vary with the matching condition?}
Table~\ref{tab:nli} presents fact-verification performance based on whether LLM-generated documents contained sentences entailed by gold evidence, focusing on GPT-4o-mini as the QE model. We observed a consistent trend across the three datasets and two expansion methods: performance measured on \emph{matched} claims (where LLM-generated documents contain sentences entailed by gold evidence) was significantly higher than that on both all and \emph{unmatched} claims, with statistical significance at \textit{p}$<$0.001. Moreover, with a few exceptions, performance on \emph{unmatched} claims was lower than that of the corresponding baseline methods without query expansion---BM25 for Query2doc and Contriever for HyDE. These trends were consistent across seven different LLMs and for $k=10$, as shown in Table~\ref{tab:app:nli_top_5} and Table~\ref{tab:app:nli_top_10}. 

\section{Discussion}

\paragraph{Query2doc and HyDE boosted fact-verification performance across three benchmarks.} 
Our findings support the effectiveness of LLM-based query expansion in zero-shot retrieval tasks~\cite{gao-etal-2023-precise,jagerman2023query, lei2024corpus, wang-etal-2024-searching}, particularly on fact-verification benchmarks. The increased verification performance through query expansion further highlights the crucial role of retrieving suitable evidence for claim verification~\cite{chen-etal-2024-complex}, suggesting a need for further research into evidence retrieval methods, such as claim decomposition~\cite{chen-etal-2022-generating,fan-etal-2020-generating}.

\paragraph{Documents generated by LLM-based query expansion methods frequently included sentences that were entailed by ground-truth evidence, indicating potential knowledge leakage.} By applying an NLI-based matching algorithm, we examined whether LLMs reproduced gold evidence in response to query expansion prompts. Our results suggest that the seven LLMs studied in this paper were likely exposed to knowledge sources from the three benchmarks during training. This observation aligns with prior research on data leakage~\cite{kandpal2023large, samuel-etal-2025-towards} and memorization in LLMs~\cite{cheng2024can, burns2022discovering}, representing the first empirical demonstration in the context of fact verification and query expansion. 

\paragraph{Performance improvements from query expansion were consistent only when LLM-generated documents contained sentences entailed by gold evidence.}
This finding suggests that the success of hypothetical document generation may be largely attributable to the LLM's internal knowledge encompassing benchmark knowledge sources. Furthermore, these results indicate that LLM-based query expansion may be limited in real-world scenarios that require retrieving niche or novel knowledge. Future research could address these limitation by incorporating external knowledge sources into the query expansion process, as demonstrated in a recent study~\cite{lei2024corpus}. 

\section{Conclusion}
This study examined the leakage hypothesis for query expansion, questioning whether an LLM generates truly hypothetical documents or merely reproduces what was exposed during the LLM's training phase. Through experiments on three widely used fact-verification benchmarks, we observed a consistent trend: on average, retrieval and verification performance increased only when the expanded query was entailed by gold evidence. This finding suggests that knowledge leakage may be present in these benchmarks, artificially inflating the perceived effectiveness of LLM-based query expansion. Since verifying claims involving niche or novel knowledge is often essential in real-world scenarios, such leakage can hinder accurate evaluation and, consequently, impede method development. These results highlight the need for future research on LLM-based query expansion methods capable of handling \emph{unknown} queries, as well as the creation of evaluation benchmarks that more accurately reflect real-world conditions involving previously unseen knowledge. 

\section*{Limitations}

While our analysis provides the evidence that certain LLM behaviors can signal memorization and that these behaviors correlate with downstream performance gains, we acknowledge three areas where further validation is needed. \textbf{(1) Causal mechanism}: Our experiments focus on LLM \emph{behaviors} rather than tracing whether specific training data directly results in the generation of associated knowledge.
Consequently, we do not claim a causal link between data leakage and generation in response to expansion prompts. Establishing such a connection remains an open challenge and would benefit from controlled studies that expose models to synthetic or novel knowledge~\cite{10.1162/tacl_a_00638, kasai2024realtime}. \textbf{(2) NLI-based matching}: To corroborate the NLI-based automatic assessment, we manually annotated a representative sample and observed a consistent trend (Table~\ref{tab:app:annotation}) with the automatic results (Table~\ref{tab:nli}). Adopting a stronger NLI model would further strengthen our conclusions. \textbf{(3) Generalizability}: Our analysis focused on three datasets within the fact-verification domain, and hence the findings are limited in scope. Further experiments are necessary to determine whether these findings generalize to other tasks, such as factual QA~\cite{joshi-etal-2017-triviaqa,10.1162/tacl_a_00276}. The methodology presented here could be adapted to those settings.

\section*{Ethics and Impact Statement}

Our findings suggest that knowledge leakage may exist in three popular fact-verification benchmarks; namely, LLMs may have been exposed to information related to the benchmarks' gold evidence during pretraining and subsequently reproduced it in response to query-expansion prompts. This observation has important implications for both fact verification and broader benchmark-oriented NLP research, as the performance of LLM-based methods may be artificially inflated due to the opaqueness of the pretraining stage. Consequently, more trustworthy evaluation frameworks and transparent LLM development practices, such as those exemplified by \citet{groeneveld-etal-2024-olmo} and \citet{gao2020pile}, are encouraged to better reflect real-world performance. We used ChatGPT to proofread portions of this manuscript.

\section*{Acknowledgments}
This research was supported by the MSIT(Ministry of Science and ICT), Korea, under the Innovative Human Resource Development for Local Intellectualization (IITP-2025-RS-2022-00156360) and Graduate School of Metaverse Convergence support (IITP-2025-RS-2024-00430997) programs, supervised by the IITP(Institute for Information \& Communications Technology Planning \& Evaluation). Kunwoo Park is the corresponding author. 

\bibliography{custom}

\appendix
\section*{Appendix}

\setcounter{figure}{0}
\setcounter{table}{0}
\renewcommand\thefigure{A\arabic{figure}} 
\renewcommand\thetable{A\arabic{table}} 

\section{Target Dataset}

This study used three fact verification benchmarks. From the BEIR benchmark~\cite{beir}, we used the test set for FEVER and the train (505) and test (188) sets for SciFact. For AVeriTeC, we used the train (3,063) and dev (500) sets, as its test set is not publicly available. We excluded claims for which gold evidence was unavailable. Table~\ref{tab:data} presents the descriptive statistics of the three datasets used in our experiments.

\section{Evaluation Metrics}
\label{sec:app:metrics}

Below, we describe the details of evaluation metric for evidence retrieval. Recall@K and NDCG@K are widely used evaluation metrics for information retrieval~\cite{manning2009introduction}, adopted in this study for evaluating evidence retrieval performance for FEVER and SciFact. Recall@K assesses the proportion of relevant items retrieved within the top K results. NDCG@K measures the quality of ranked results by considering both the relevance of retrieved documents and their positions within the ranking. We used pyrec\_eval~\cite{VanGysel2018pytreceval} to measure Recall@K and NDCG@K.

\begin{equation}
\label{eq:hungarian}
    \begin{aligned}
        S(\hat{Y}, Y) &= \frac{1}{|Y|}\text{max}\sum_{\hat{y}\in \hat{Y}}\sum_{y\in Y}f(\hat{y}, y)X(\hat{y},y) 
    \end{aligned}
\end{equation}

For AVeriTeC, where gold evidence is not selected from a knowledge store but written by human annotators, we used METEOR and BERTScore with the Hungarian algorithm. Equation~\ref{eq:hungarian} presents the algorithm, where $f$ is a pairwise scoring function, and $X$ is a boolean function representing the assignment between the generated sequences $\hat{Y}$ and the reference sequences $Y$. To measure token-level and embedding-level similarity, we used METEOR and BERTScore for $f$, respectively. METEOR was computed using NLTK~\cite{bird2009natural}, and BERTScore was calculated with DeBERTa-xlarge-MNLI\footnote{\url{https://huggingface.co/microsoft/deberta-xlarge-mnli}}. We reported observations for varying $k$: Table~\ref{tab:fact_verification_top_5} and Table~\ref{tab:nli} for $k=5$, and Table~\ref{tab:app:fact_verification_top_10} and Table~\ref{tab:app:nli_top_10} for $k=10$.

\section{Experimental Setups}
For HyDE, we encoded queries and documents using Contriever\footnote{\url{https://huggingface.co/facebook/contriever}}. For Query2doc, we used BM25 provided in PySerini~\cite{Lin_etal_SIGIR2021_Pyserini}. Following the same settings in previous study~\cite{gao-etal-2023-precise}, we set LLM hyperparameters as follows: temperature as 0.7, top\_p as 1.0, and max\_tokens as 512. We used the Mann–Whitney 
U test for statistical testing on performance differences.

\section{Computing Environment}
We ran experiments using two machines. The first machine has four Nvidia RTX A6000 GPUs (48GB per GPU) and 256GB RAM. The second machine has two Nvidia H100 GPUs (80GB per GPU) and 480GB RAM. The experiments were conducted in a computing environment with the following configuration: Python 3.11.10, PyTorch 2.3.1, Transformers 4.43.4, vLLM 0.5.3, pyrec-eval 0.5, Faiss 1.8, Pyserini 0.40.0, NLTK 3.9.1, bert-score 0.3.13, rouge-score 0.1.2.

We used GPT-4o-mini, Claude 3 Haiku, Gemini 1.5 Flash via API, while Llama 3.1 (8B and 70B) and Mistral 7B, and Mixtral 8x7B were accessed through pretrained checkpoints. The model ids and parameter sizes are provided below.

\begin{itemize}
    \item GPT-4o-mini: \texttt{gpt-4o-mini-2024-07-18} (Parameter size: unknown)
    \item Claude-3-haiku: \texttt{claude-3-haiku-20240307} (Parameter size: unknown)
    \item Gemini-1.5-flash: \texttt{gemini-1.5-flash} (Parameter size: unknown)
    \item Llama-3.1-8b-it: \url{https://huggingface.co/meta-llama/Llama-3.1-8B-Instruct} (Parameter size: 8B)
    \item Llama-3.1-70b-it: \url{https://huggingface.co/meta-llama/Llama-3.1-70B-Instruct} (Parameter size: 70B)
    \item Mistral-7b-it: \url{https://huggingface.co/mistralai/Mistral-7B-Instruct-v0.3} (Parameter size: 7B)
    \item Mixtral-8x7b-it: \url{https://huggingface.co/mistralai/Mixtral-8x7B-Instruct-v0.1} (Parameter size: 46.7B)
\end{itemize}

\begin{figure}[t]
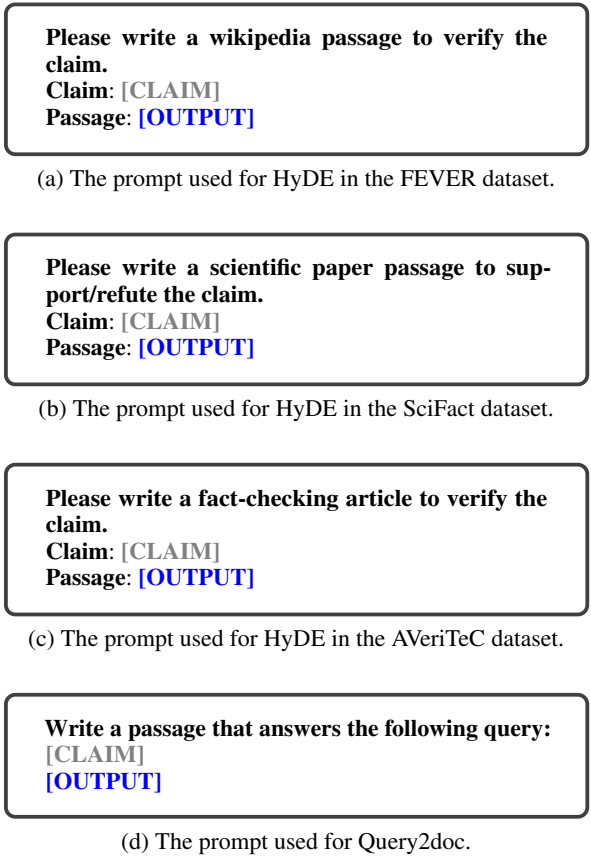

\begin{subfigure}[b]{\linewidth}
\begin{tcolorbox}[colback=white, fontupper=\small]
\textbf{Please write a wikipedia passage to verify the claim.}\\
\textbf{Claim}: \textcolor{gray}{\textbf{[CLAIM]}} \\
\textbf{Passage}: \textcolor{blue}{\textbf{[OUTPUT]}}
\end{tcolorbox}
\centering
\vspace*{-3mm}
\caption{The prompt used for HyDE in the FEVER dataset.}
\label{fig:app:fev_hyde_prompt}
\end{subfigure}

\hfill

\begin{subfigure}[b]{\linewidth}
\begin{tcolorbox}[colback=white, fontupper=\small]
\textbf{Please write a scientific paper passage to support/refute the claim.}\\
\textbf{Claim}: \textcolor{gray}{\textbf{[CLAIM]}} \\
\textbf{Passage}: \textcolor{blue}{\textbf{[OUTPUT]}}
\end{tcolorbox}
\centering
\vspace*{-3mm}
\caption{The prompt used for HyDE in the SciFact dataset.}
\label{fig:app:sci_hyde_prompt}
\end{subfigure}

\hfill

\begin{subfigure}[b]{\linewidth}
\begin{tcolorbox}[colback=white, fontupper=\small]
\textbf{Please write a fact-checking article to verify the claim.}\\
\textbf{Claim}: \textcolor{gray}{\textbf{[CLAIM]}} \\
\textbf{Passage}: \textcolor{blue}{\textbf{[OUTPUT]}}
\end{tcolorbox}
\centering
\vspace*{-3mm}
\caption{The prompt used for HyDE in the AVeriTeC dataset.}
\label{fig:app:aver_hyde_prompt}
\end{subfigure}

\hfill

\begin{subfigure}[b]{\linewidth}
\begin{tcolorbox}[colback=white, fontupper=\small]
\textbf{Write a passage that answers the following query:} \textcolor{gray}{\textbf{[CLAIM]}}\\ \textcolor{blue}{\textbf{[OUTPUT]}}
\end{tcolorbox}
\centering
\vspace*{-3mm}
\caption{The prompt used for Query2doc.}
\label{fig:app:q2d_prompt}
\end{subfigure}
\vspace*{-4mm}
\caption{Prompts used for query expansion.}
\label{fig:app:qe_prompt}
\end{figure}

\section{Prompts}
\label{app:prompt}

\paragraph{Query Expansion} Figure~\ref{fig:app:fev_hyde_prompt},~\ref{fig:app:sci_hyde_prompt}, and ~\ref{fig:app:aver_hyde_prompt} illustrate the HyDE prompts used in this study where the original prompt is adapted to each dataset. For Query2doc, we used the same prompt by following the suggestion in \citet{wang-etal-2023-query2doc}, as shown in Figure~\ref{fig:app:q2d_prompt}.

\begin{figure}[t]
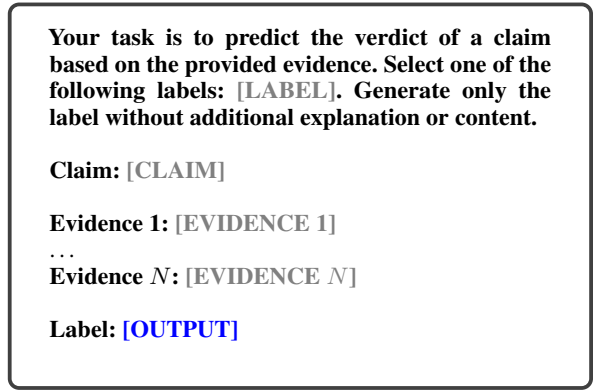

\begin{tcolorbox}[colback=white, fontupper=\small]
\textbf{Your task is to predict the verdict of a claim based on the provided evidence. Select one of the following labels:} \textcolor{gray}{\textbf{[LABEL]}}\textbf{.
Generate only the label without additional explanation or content.} \\ \\
\textbf{Claim:} \textcolor{gray}{\textbf{[CLAIM]}} \\\\
\textbf{Evidence 1:} \textcolor{gray}{\textbf{[EVIDENCE 1]}} \\
…\\
\textbf{Evidence $N$:} \textcolor{gray}{\textbf{[EVIDENCE $N$]}} \\\\
\textbf{Label: }\textcolor{blue}{\textbf{[OUTPUT]}}\\
\end{tcolorbox}
\centering
\vspace*{-4mm}
\caption{The prompt used for fact verification with GPT-4o-mini. $N$ denotes the total number of retrieved evidence.}
\label{fig:app:verification_prompt}
\end{figure}

\paragraph{Verdict Prediction} Figure~\ref{fig:app:verification_prompt} presents the prompt used for verdict prediction with GPT-4o-mini.

\begin{figure}[t]
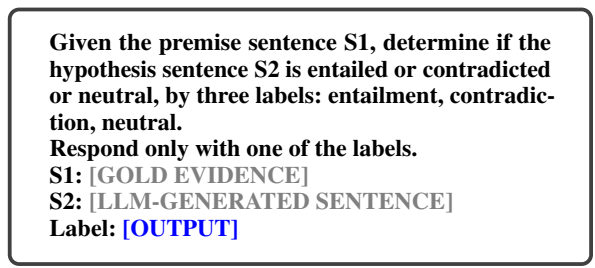

\begin{tcolorbox}[colback=white, fontupper=\small]
\textbf{Given the premise sentence S1, determine if the hypothesis sentence S2 is entailed or contradicted or neutral, by three labels: entailment, contradiction, neutral.\\
Respond only with one of the labels. \\
S1: \textcolor{gray}{\textbf{[GOLD EVIDENCE]}} \\
S2: \textcolor{gray}{\textbf{[LLM-GENERATED SENTENCE]}} \\
Label: }\textcolor{blue}{\textbf{[OUTPUT]}}
\end{tcolorbox}
\centering
\vspace*{-4mm}
\caption{The prompt used for NLI.}
\label{fig:app:NLI_prompt}
\end{figure}

\paragraph{Natural Language Inference} 
We used GPT-4o-mini for natural language inference, employing a prompt proposed in a previous study~\cite{wang-etal-2024-rethinking}, as illustrated in Figure~\ref{fig:app:NLI_prompt}. To support its validity, two authors manually annotated the labels for a randomly selected set of 200 pairs following the guidelines presented in Figure~\ref{fig:app:nli_guideline}. The GPT-based NLI model achieved an F1 score of 0.8 on the sampled data.

\section{Supplementary Results}
\label{sec:app:results}

\paragraph{LLM comparison for fact verification} Table~\ref{tab:fact_verification_top_5} present the results for evidence retrieval and verdict prediction by varying backbone LLMs for query expansion. We observed that Llama-3.1-70b-it generally performed well when used with Query2doc. For HyDE, while Llama-3.1-70b-it achieved the best performance on FEVER, Claude-3-haiku obtained higher evaluation scores in SciFact and AVeriTeC.

\paragraph{Proportion of matched claims across different benchmarks}
Table~\ref{tab:matched} presents the distribution of matched claims across three datasets, varying the LLMs used for query expansion. On average, the estimated proportion was higher for FEVER than for the other two datasets. While FEVER was constructed using public Wikipedia documents, SciFact is based on scientific literature, covering \textit{niche} knowledge, and AVeriTeC is the most recent dataset based on web documents collected by human annotators, covering \textit{recent} knowledge. Given these characteristics, the highest proportion observed in FEVER partially supports the validity of the NLI-based estimation.

\begin{figure}[t]
\centering
\begin{tcolorbox}[
    colback=white,
    colframe=black,
    fonttitle=\bfseries,
    fontupper=\small,
    sharp corners
]
\textbf{Premise:} \textcolor{gray}{\textbf{[GOLD EVIDENCE]}} \\
\textbf{Hypothesis:} \\ 
\textcolor{gray}{\textbf{[LLM-GENERATED SENTENCE]}} \\

\textbf{Given the premise, determine whether the hypothesis is entailed.}
\begin{itemize}
    \item[$\square$] \textbf{Entailment}
    \item[$\square$] \textbf{Non-Entailment}
\end{itemize}
\end{tcolorbox}
\vspace{-4mm}
\caption{Manual labeling guidelines for natural language inference.}
\vspace{-4mm}
\label{fig:app:nli_guideline}
\end{figure}

\begin{figure}[t]
\centering
\begin{tcolorbox}[
    colback=white,
    colframe=black,
    fonttitle=\bfseries,
    fontupper=\small,
    sharp corners
]
\textbf{Claim:} \textcolor{gray}{\textbf{[CLAIM]}} \\
\textbf{Gold Evidence:} \textcolor{gray}{\textbf{[GOLD EVIDENCE]}} \\
\textbf{LLM-generated Document:} \\ \textcolor{gray}{\textbf{[LLM-GENERATED DOCUMENT]}} \\

\textbf{Determine whether the LLM-generated document contains the whole or part of any gold evidence.}
\begin{itemize}
    \item[$\square$] \textbf{Included}
    \item[$\square$] \textbf{Not Included}
\end{itemize}
\end{tcolorbox}
\vspace{-4mm}
\caption{Manual labeling guidelines for determining whether LLM-generated documents contain gold evidence.}
\vspace{-4mm}
\label{fig:app:matched_guideline}
\end{figure}

\paragraph{Effects of retrieving more evidence} Table~\ref{tab:app:fact_verification_top_10} presents the fact verification performance with an increased number of retrieved evidence ($k=10$). Performance improves in every case across different LLMs and datasets compared to the results with $k=5$. Table~\ref{tab:app:nli_top_10} shows performance depending on the matching condition, showing a consistent trend with Table~\ref{tab:app:nli_top_5} when retrieving more evidence.

\begin{table*}[ht]
\centering
\begin{subtable}{0.99\textwidth}
    \centering
    \resizebox{\textwidth}{!}{
    \begin{tabular}{c|ccc|ccc|ccc}
    \hline
    \multirow{2}{*}{Method} & \multicolumn{3}{c|}{FEVER} & \multicolumn{3}{c|}{SciFact} & \multicolumn{3}{c}{AVeriTeC} \\ \cline{2-10}
    & Recall@10 & NDCG@10 & F1 & Recall@10 & NDCG@10 & F1 & METEOR & BERTScore & F1\\ \hline
    BM25        & 37.1  & 27.1  & 55.3  & 58.4  & 48.3  & 50.6  & 21   & 15    & 33.1 \\\hline 
    \multicolumn{10}{c}{Performance by varying LLMs}\\\hline
    GPT-4o-mini      & 44.2$\pm$0.1  & 32$\pm$0.1  & 57$\pm$0.1  & 64$\pm$0.0  & 51.4$\pm$0.1  & \textbf{53.2$\pm$0.6}  & 22.3$\pm$0.0  & 15.9$\pm$0.0  & 34.6$\pm$0.1 \\
    Claude-3-haiku   & 43.4$\pm$0.1  & 31$\pm$0.1  & 56.9$\pm$0.2  & 64.3$\pm$0.2  & 51.5$\pm$0.1  & \textbf{53$\pm$0.5} & 22.5$\pm$0.0  & 16$\pm$0.0    & 34.7$\pm$0.1 \\
    Gemini-1.5-flash & 43.4$\pm$0.1  & 31.7$\pm$0.0  & 56.9$\pm$0.1  & \textbf{64.7$\pm$0.2}  & \textbf{52.8$\pm$0.1}  & \textbf{53.1$\pm$0.6} & 22.3$\pm$0.0  & 16.2$\pm$0.0  & \textbf{34.7$\pm$0.2}  \\\hline
    Llama-3.1-8b-it  & 43.6$\pm$0.1  & 31.3$\pm$0.1  & 56.8$\pm$0.1  & 63.2$\pm$0.3  & 51.2$\pm$0.1  & \textbf{53$\pm$0.5}  & 22.3$\pm$0.0  & 15.9$\pm$0.0  & 34.5$\pm$0.1   \\
    Llama-3.1-70b-it & \textbf{46.1$\pm$0.2}  & \textbf{33.7$\pm$0.1}  & \textbf{57.2$\pm$0.2}  & \textbf{64.6$\pm$0.2}   & 52.5$\pm$0.1  & \textbf{53.1$\pm$0.3} & \textbf{22.7$\pm$0.1}  & \textbf{16.3$\pm$0.0}  & \textbf{34.8$\pm$0.1}   \\
    Mistral-7b-it    & 43.2$\pm$0.2  & 30.8$\pm$0.2  & 56.8$\pm$0.1  & 63.1$\pm$0.2  & 51$\pm$0.1    & 52.6$\pm$0.4 & 22.5$\pm$0.0  & 16.2$\pm$0.0  & 34.6$\pm$0.1   \\
    Mixtral-8x7b-it  & 43.4$\pm$0.2  & 30.8$\pm$0.2  & 56.8$\pm$0.2  & 63.6$\pm$0.2    & 51.3$\pm$0.1  & \textbf{52.8$\pm$0.4} & 22.6$\pm$0.0  & 16.1$\pm$0.0  & \textbf{34.9$\pm$0.1}   \\ \hline
    \end{tabular}
    }
    \caption{Query2doc}
    \vspace{2mm}
\end{subtable}
\hfill
\begin{subtable}{0.99\textwidth}
    \centering
    \resizebox{\textwidth}{!}{
    \begin{tabular}{c|ccc|ccc|ccc}
    \hline
    \multirow{2}{*}{Method} & \multicolumn{3}{c|}{FEVER} & \multicolumn{3}{c|}{SciFact} & \multicolumn{3}{c}{AVeriTeC} \\ \cline{2-10}
    & Recall@10 & NDCG@10 & F1 & Recall@10 & NDCG@10 & F1 & METEOR & BERTScore & F1\\ \hline
    Contriever  & 34.4  & 22.8  & 54.8  & 65    & 51.2  & 54  & 20.8 & 16.1  & 34.8    \\ \hline
    \multicolumn{10}{c}{Performance by varying LLMs}\\\hline
    GPT-4o-mini      & 46.7$\pm$0.1 & 32.1$\pm$0.0  & 56.7$\pm$0.1  & 70$\pm$0.1    & 56.7$\pm$0.1  & 54.6$\pm$0.3  & 22.3$\pm$0.0  & 17$\pm$0.0    & 36.6$\pm$0.4 \\
    Claude-3-haiku   & 46.2$\pm$0.0 & 31.4$\pm$0.0  & 56.7$\pm$0.1  & \textbf{71.6$\pm$0.2}  & \textbf{58.3$\pm$0.1}  & \textbf{55$\pm$0.3}  & \textbf{22.8$\pm$0.0}  & \textbf{17.8$\pm$0.0}  & \textbf{37.6$\pm$0.3}  \\
    Gemini-1.5-flash & 44.2$\pm$0.1 & 29.9$\pm$0.1  & 56.5$\pm$0.1  & 69.8$\pm$0.1  & 56.6$\pm$0.2  & \textbf{54.8$\pm$0.3} & 21.6$\pm$0.0  & 16.3$\pm$0.0  & \textbf{37.1$\pm$0.8}  \\ \hline
    Llama-3.1-8b-it  & 46.4$\pm$0.1 & 31.8$\pm$0.1  & 56.6$\pm$0.1  & 70.1$\pm$0.2  & 57$\pm$0.1    & 54.3$\pm$0.5  & 22.4$\pm$0.0  & 17.4$\pm$0.0  & 36.8$\pm$0.3 \\
    Llama-3.1-70b-it & \textbf{49.7$\pm$0.2} & \textbf{35$\pm$0.2}  & \textbf{57$\pm$0.1} & 70.8$\pm$0.2  & 57.8$\pm$0.2  & \textbf{55$\pm$0.5} & 22.5$\pm$0.2  & 17.5$\pm$0.1  & \textbf{36.7$\pm$0.9}  \\
    Mistral-7b-it    & 46.1$\pm$0.1 & 31.2$\pm$0.0  & 56.5$\pm$0.1  & 69.8$\pm$0.2  & 56.4$\pm$0.2  & \textbf{54.8$\pm$0.3} & 22.7$\pm$0.0  & 17.5$\pm$0.0  & \textbf{37.3$\pm$0.5}  \\
    Mixtral-8x7b-it  & 47.6$\pm$0.0 & 32.6$\pm$0.0  & 56.9$\pm$0.2 & 70.2$\pm$0.2  & 56.8$\pm$0.1  & \textbf{54.8$\pm$0.5} & \textbf{22.8$\pm$0.0}  & 17.6$\pm$0.0  & \textbf{37.2$\pm$0.5}  \\ \hline
    \end{tabular}
    }
    \caption{HyDE}
\end{subtable}
\caption{Fact verification performance using baseline retrievers and LLM-based query expansion methods, with the number of retrieved evidence set to ten ($k=10$). We report the average performance of query expansion methods along with standard errors, obtained by repeating the generations eight times.}
\label{tab:app:fact_verification_top_10}
\end{table*}

\begin{table*}[ht]
\small
\centering
\hfill
\begin{subtable}{0.99\textwidth}
    \centering
    \resizebox{\textwidth}{!}{
    \begin{tabular}{cc|ccc|ccc|ccc}
    \hline
    \multirow{2}{*}{Method} & \multirow{2}{*}{Data} & \multicolumn{3}{c|}{FEVER} & \multicolumn{3}{c|}{SciFact} & \multicolumn{3}{c}{AVeriTeC}  \\ \cline{3-11}
    && Recall@5 & NDCG@5 & F1 & Recall@5 & NDCG@5 & F1 & METEOR & BERTScore & F1\\ \hline
    \multirow{3}{*}{Query2doc} & ALL  & 35.2$\pm$0.1 & 28.3$\pm$0.1 & 55.4$\pm$0.1 & 56  $\pm$0.1 & 48.2$\pm$0.1 & 52$\pm$0.5 & 19.3$\pm$0.0 & 12.5$\pm$0.0 & 33.1$\pm$0.1  \\
    & $M$                             & \textbf{38.7$\pm$0.1} & \textbf{31.2$\pm$0.1} & \textbf{58$\pm$0.0} & \textbf{63.7$\pm$0.3} & \textbf{56.7$\pm$0.3} & \textbf{54.1$\pm$0.2} & \textbf{22$\pm$0.0} & \textbf{17.7$\pm$0.0} & \textbf{38.8$\pm$0.3}  \\
    & $\neg M$                        & 23.8$\pm$0.1 & 18.5$\pm$0.1 & 44.7$\pm$0.1 & 46.2$\pm$0.6 & 37.2$\pm$0.4 & 48$\pm$0.5 & 17.4$\pm$0.0 & 8.9$\pm$0.0 & 27.6$\pm$0.1  \\ \hline
    \multirow{3}{*}{HyDE} & ALL  & 36.7$\pm$0.1 & 28.1$\pm$0.0 & 55.6$\pm$0.1 & 62.8$\pm$0.1 & 54.7$\pm$0.1 & 53.7$\pm$0.4 & 19.3$\pm$0.0 & 14  $\pm$0.0 & 36.2$\pm$0.6  \\
    & $M$                        & \textbf{39.9$\pm$0.2} & \textbf{30.7$\pm$0.1} & \textbf{57.5$\pm$0.1} & \textbf{71.1$\pm$0.5} & \textbf{63.8$\pm$0.4} & \textbf{57.5$\pm$0.3} & \textbf{20.4$\pm$0.1} & \textbf{16.5$\pm$0.1} & \textbf{37.8$\pm$0.3}  \\
    & $\neg M$                   & 25.5$\pm$0.2 & 18.9$\pm$0.2 & 45$\pm$0.2 & 53$\pm$0.5 & 44.1$\pm$0.5 & 48.8$\pm$0.4 & 17.3$\pm$0.1 & 9.8$\pm$0.1 & 31.9$\pm$0.3  \\\hline
    \end{tabular}}
    \caption{Claude-3-haiku}
    \vspace{2mm}
\end{subtable}
\hfill
\begin{subtable}{0.99\textwidth}
    \centering
    \resizebox{\textwidth}{!}{
    \begin{tabular}{cc|ccc|ccc|ccc}
    \hline
    \multirow{2}{*}{Method} & \multirow{2}{*}{Data} & \multicolumn{3}{c|}{FEVER} & \multicolumn{3}{c|}{SciFact} & \multicolumn{3}{c}{AVeriTeC}  \\ \cline{3-11}
    && Recall@5 & NDCG@5 & F1 & Recall@5 & NDCG@5 & F1 & METEOR & BERTScore & F1\\ \hline
    \multirow{3}{*}{Query2doc} & ALL  & 36.2$\pm$0.1 & 29.2$\pm$0.1 & 55.8$\pm$0.1 & 56.2$\pm$0.2 & 49.4$\pm$0.1 & 52.2$\pm$0.5 & 18.9$\pm$0.0 & 12.5$\pm$0.0 & 33.3$\pm$0.2  \\
    & $M$                             & \textbf{38.5$\pm$0.1} & \textbf{31.4$\pm$0.0} & \textbf{58.8$\pm$0.0} & \textbf{63.4$\pm$0.5} & \textbf{56.9$\pm$0.4} & \textbf{55$\pm$0.5} & \textbf{20.5$\pm$0.0} & \textbf{16.5$\pm$0.1} & \textbf{38.1$\pm$0.4}  \\
    & $\neg M$                        & 30.7$\pm$0.1 & 24.1$\pm$0.1 & 48$\pm$0.2 & 48.9$\pm$0.4 & 41.6$\pm$0.3 & 48.7$\pm$0.3 & 17.6$\pm$0.0 & 9.4$\pm$0.1 & 28.5$\pm$0.2  \\ \hline
    \multirow{3}{*}{HyDE} & ALL  & 35$\pm$0.1 & 26.7$\pm$0.1 & 55.2$\pm$0.2 & 61$\pm$0.2 & 52.9$\pm$0.2 & 53.5$\pm$0.7 & 18$\pm$0.0 & 12.4$\pm$0.0 & 35.7$\pm$0.5  \\
    & $M$                        & \textbf{37.1$\pm$0.1} & \textbf{28.2$\pm$0.1} & \textbf{59$\pm$0.1} & \textbf{67.1$\pm$0.6} & \textbf{60.1$\pm$0.4} & \textbf{57.3$\pm$0.5} & \textbf{18.8$\pm$0.0} & \textbf{14.7$\pm$0.1} & \textbf{37.3$\pm$0.2}  \\
    & $\neg M$                   & 30.1$\pm$0.1 & 23.1$\pm$0.1 & 45$\pm$0.2 & 58.7$\pm$0.4 & 50.2$\pm$0.2 & 52$\pm$0.3 & 16.8$\pm$0.1 & 9.2$\pm$0.1 & 31.6$\pm$0.3  \\\hline
    \end{tabular}}
    \caption{Gemini-1.5-flash}
    \vspace{2mm}
\end{subtable}
\hfill
\begin{subtable}{0.99\textwidth}
    \centering
    \resizebox{\textwidth}{!}{
    \begin{tabular}{cc|ccc|ccc|ccc}
    \hline
    \multirow{2}{*}{Method} & \multirow{2}{*}{Data} & \multicolumn{3}{c|}{FEVER} & \multicolumn{3}{c|}{SciFact} & \multicolumn{3}{c}{AVeriTeC}  \\ \cline{3-11}
    && Recall@5 & NDCG@5 & F1 & Recall@5 & NDCG@5 & F1 & METEOR & BERTScore & F1\\ \hline
    \multirow{3}{*}{Query2doc} & ALL  & 35.7$\pm$0.1 & 28.6$\pm$0.2 & 55.6$\pm$0.2 & 54.9$\pm$0.2 & 47.8$\pm$0.2 & 51.9$\pm$0.3 & 19$\pm$0.0 & 12.4$\pm$0.0 & 32.2$\pm$0.2  \\
    & $M$                             & \textbf{39$\pm$0.2} & \textbf{31.4$\pm$0.2} & \textbf{58.6$\pm$0.1} & \textbf{63.9$\pm$0.4} & \textbf{57.5$\pm$0.4} & \textbf{55.1$\pm$0.4} & \textbf{21.7$\pm$0.1} & \textbf{17.3$\pm$0.1} & \textbf{36.6$\pm$0.3}  \\
    & $\neg M$                        & 28.5$\pm$0.3 & 22.3$\pm$0.3 & 48.5$\pm$0.3 & 44.2$\pm$0.6 & 36.5$\pm$0.4 & 47.6$\pm$0.3 & 17.4$\pm$0.0 & 9.4$\pm$0.1 & 28.5$\pm$0.1  \\ \hline
    \multirow{3}{*}{HyDE} & ALL  & 36.7$\pm$0.1 & 28.4$\pm$0.1 & 55.4$\pm$0.1 & 61.2$\pm$0.2 & 53.4$\pm$0.2 & 53.6$\pm$0.7 & 18.9$\pm$0.0 & 13.6$\pm$0.0 & 35.5$\pm$0.4  \\
    & $M$                        & \textbf{39.5$\pm$0.1} & \textbf{30.6$\pm$0.1} & \textbf{58.2$\pm$0.1} & \textbf{68.7$\pm$0.7} & \textbf{62.3$\pm$0.6} & \textbf{56.3$\pm$0.4} & \textbf{20$\pm$0.1} & \textbf{16.1$\pm$0.1} & \textbf{37.7$\pm$0.2}  \\
    & $\neg M$                   & 29.2$\pm$0.2 & 22.4$\pm$0.2 & 46.8$\pm$0.2 & 54.1$\pm$0.8 & 45.1$\pm$0.6 & 50.6$\pm$0.4 & 17.7$\pm$0.1 & 10.6$\pm$0.1 & 31.8$\pm$0.3  \\\hline
    \end{tabular}}
    \caption{Llama-3.1-8b-it}
    \vspace{2mm}
\end{subtable}
\hfill
\begin{subtable}{0.99\textwidth}
    \centering
    \resizebox{\textwidth}{!}{
    \begin{tabular}{cc|ccc|ccc|ccc}
    \hline
    \multirow{2}{*}{Method} & \multirow{2}{*}{Data} & \multicolumn{3}{c|}{FEVER} & \multicolumn{3}{c|}{SciFact} & \multicolumn{3}{c}{AVeriTeC}  \\ \cline{3-11}
    && Recall@5 & NDCG@5 & F1 & Recall@5 & NDCG@5 & F1 & METEOR & BERTScore & F1\\ \hline
    \multirow{3}{*}{Query2doc} & ALL  & 38.3$\pm$0.1 & 31$\pm$0.1 & 56.1$\pm$0.1 & 56.4$\pm$0.3 & 49.2$\pm$0.1 & 52.4$\pm$0.7 & 19.3$\pm$0.1 & 12.7$\pm$0.0 & 33.4$\pm$0.2  \\
    & $M$                             & \textbf{41.3$\pm$0.1} & \textbf{33.6$\pm$0.1} & \textbf{58.6$\pm$0.1} & \textbf{65$\pm$0.3} & \textbf{58.3$\pm$0.2} & \textbf{54.9$\pm$0.5} & \textbf{21.6$\pm$0.1} & \textbf{17.2$\pm$0.1} & \textbf{38.1$\pm$0.3}  \\
    & $\neg M$                        & 27.6$\pm$0.4 & 21.7$\pm$0.4 & 45.9$\pm$0.3 & 44.9$\pm$0.5 & 37$\pm$0.3 & 47.9$\pm$0.2 & 17.3$\pm$0.1 & 8.6$\pm$0.1 & 27.6$\pm$0.2  \\ \hline
    \multirow{3}{*}{HyDE} & ALL  & 40.4$\pm$0.2 & 31.7$\pm$0.2 & 55.9$\pm$0.1 & 61.9$\pm$0.3 & 54.1$\pm$0.2 & 53.6$\pm$0.5 & 19$\pm$0.2 & 13.7$\pm$0.1 & 35.4$\pm$0.7  \\
    & $M$                        & \textbf{44.3$\pm$0.5} & \textbf{35$\pm$0.5} & \textbf{58.4$\pm$0.1} & \textbf{69.2$\pm$0.4} & \textbf{62.1$\pm$0.4} & \textbf{56.7$\pm$0.3} & \textbf{20.9$\pm$0.2} & \textbf{17.3$\pm$0.3} & \textbf{38.3$\pm$0.2}  \\
    & $\neg M$                   & 30.4$\pm$0.3 & 23.3$\pm$0.3 & 48.9$\pm$0.2 & 52.9$\pm$0.7 & 44.3$\pm$0.6 & 48.9$\pm$0.5 & 17.4$\pm$0.1 & 10.6$\pm$0.2 & 31.5$\pm$0.4  \\\hline
    \end{tabular}}
    \caption{Llama-3.1-70b-it}
    \vspace{2mm}
\end{subtable}
\hfill
\begin{subtable}{0.99\textwidth}
    \centering
    \resizebox{\textwidth}{!}{
    \begin{tabular}{cc|ccc|ccc|ccc}
    \hline
    \multirow{2}{*}{Method} & \multirow{2}{*}{Data} & \multicolumn{3}{c|}{FEVER} & \multicolumn{3}{c|}{SciFact} & \multicolumn{3}{c}{AVeriTeC}  \\ \cline{3-11}
    && Recall@5 & NDCG@5 & F1 & Recall@5 & NDCG@5 & F1 & METEOR & BERTScore & F1\\ \hline
    \multirow{3}{*}{Query2doc} & ALL  & 35.1$\pm$0.3 & 28$\pm$0.2 & 55.4$\pm$0.2 & 55.1$\pm$0.1 & 47.9$\pm$0.1 & 51.9$\pm$0.6 & 19.2$\pm$0.0 & 12.6$\pm$0.0 & 32.8$\pm$0.1  \\
    & $M$                             & \textbf{39$\pm$0.2} & \textbf{31.3$\pm$0.2} & \textbf{58.2$\pm$0.1} & \textbf{64.8$\pm$0.3} & \textbf{58$\pm$0.2} & \textbf{54.8$\pm$0.5} & \textbf{21.5$\pm$0.1} & \textbf{17.3$\pm$0.1} & \textbf{37.5$\pm$0.5}  \\
    & $\neg M$                        & 24.8$\pm$0.4 & 19.1$\pm$0.3 & 46.7$\pm$0.2 & 44.9$\pm$0.5 & 37.2$\pm$0.3 & 48.3$\pm$0.3 & 17.4$\pm$0.0 & 8.8$\pm$0.1 & 27.7$\pm$0.2  \\ \hline
    \multirow{3}{*}{HyDE} & ALL  & 36.3$\pm$0.1 & 27.8$\pm$0.0 & 55.3$\pm$0.1 & 60.7$\pm$0.2 & 52.7$\pm$0.2 & 53.4$\pm$0.4 & 19$\pm$0.0 & 13.6$\pm$0.0 & 35.8$\pm$0.7  \\
    & $M$                        & \textbf{39.2$\pm$0.2} & \textbf{30.1$\pm$0.1} & \textbf{57.7$\pm$0.1} & \textbf{67.8$\pm$0.5} & \textbf{60.9$\pm$0.4} & \textbf{55.5$\pm$0.7} & \textbf{20.1$\pm$0.1} & \textbf{16$\pm$0.1} & \textbf{36.8$\pm$0.4}  \\
    & $\neg M$                   & 27.5$\pm$0.2 & 20.9$\pm$0.1 & 46.4$\pm$0.2 & 53.8$\pm$0.3 & 44.8$\pm$0.3 & 50.4$\pm$0.3 & 17.7$\pm$0.1 & 10.6$\pm$0.1 & 33.1$\pm$0.4  \\\hline
    \end{tabular}}
    \caption{Mistral-7b-it}
    \vspace{2mm}
\end{subtable}
\hfill
\begin{subtable}{0.99\textwidth}
    \centering
    \resizebox{\textwidth}{!}{
    \begin{tabular}{cc|ccc|ccc|ccc}
    \hline
    \multirow{2}{*}{Method} & \multirow{2}{*}{Data} & \multicolumn{3}{c|}{FEVER} & \multicolumn{3}{c|}{SciFact} & \multicolumn{3}{c}{AVeriTeC}  \\ \cline{3-11}
    && Recall@5 & NDCG@5 & F1 & Recall@5 & NDCG@5 & F1 & METEOR & BERTScore & F1\\ \hline
    \multirow{3}{*}{Query2doc} & ALL  & 35.1$\pm$0.2 & 27.9$\pm$0.2 & 55.3$\pm$0.2 & 54.6$\pm$0.2 & 47.7$\pm$0.1 & 51.9$\pm$0.4 & 19.4$\pm$0.0 & 12.7$\pm$0.0 & 33.2$\pm$0.1  \\
    & $M$                             & \textbf{38.6$\pm$0.2} & \textbf{30.9$\pm$0.2} & \textbf{57.8$\pm$0.1} & \textbf{63.5$\pm$0.5} & \textbf{57.3$\pm$0.4} & \textbf{54.1$\pm$0.3} & \textbf{21.5$\pm$0.1} & \textbf{17$\pm$0.1} & \textbf{37.3$\pm$0.3}  \\
    & $\neg M$                        & 22.3$\pm$0.3 & 17.1$\pm$0.2 & 44.3$\pm$0.2 & 43.4$\pm$0.4 & 35.6$\pm$0.3 & 47.8$\pm$0.3 & 17.3$\pm$0.1 & 8.4$\pm$0.1 & 27.7$\pm$0.3  \\ \hline
    \multirow{3}{*}{HyDE} & ALL  & 37.6$\pm$0.1 & 29.1$\pm$0.1 & 55.7$\pm$0.1 & 61.3$\pm$0.2 & 53.1$\pm$0.1 & 53.3$\pm$0.3 & 19.2$\pm$0.0 & 13.7$\pm$0.0 & 35.8$\pm$0.7  \\
    & $M$                        & \textbf{40.3$\pm$0.1} & \textbf{31.2$\pm$0.1} & \textbf{57.7$\pm$0.0} & \textbf{68.9$\pm$0.4} & \textbf{61.4$\pm$0.3} & \textbf{55.4$\pm$0.2} & \textbf{20.3$\pm$0.1} & \textbf{16.1$\pm$0.1} & \textbf{37.1$\pm$0.2}  \\
    & $\neg M$                   & 26.4$\pm$0.2 & 20.4$\pm$0.2 & 45.2$\pm$0.2 & 52.2$\pm$0.3 & 43.1$\pm$0.2 & 49.7$\pm$0.4 & 17.7$\pm$0.1 & 10.6$\pm$0.1 & 32.6$\pm$0.4  \\\hline
    \end{tabular}}
    \caption{Mixtral-8x7b-it}
    \vspace{2mm}
\end{subtable}
\vspace{-2mm}
\caption{Fact verification performance depending on whether the document generated by query expansion methods contains sentences entailed by gold evidence, with the number of retrieved evidence set to five ($k=5$). We report performance using different backbone LLMs for query expansion.}
\label{tab:app:nli_top_5}
\end{table*}

\begin{table*}[ht]
\small
\centering
\hfill
\begin{subtable}{\textwidth}
    \centering
    \resizebox{.99\textwidth}{!}{
    \begin{tabular}{cc|ccc|ccc|ccc}
    \hline
    \multirow{2}{*}{Method} &\multirow{2}{*}{Data} & \multicolumn{3}{c|}{FEVER} & \multicolumn{3}{c|}{SciFact} & \multicolumn{3}{c}{AVeriTeC} \\ \cline{3-11}
    & & Recall@10 & NDCG@10 & F1 & Recall@10 & NDCG@10 & F1 & METEOR & BERTScore & F1\\ \hline
    \multirow{3}{*}{Query2doc} & ALL               & 44.2$\pm$0.1 & 32$\pm$0.1 & 57$\pm$0.1 & 64$\pm$0.0 & 51.4$\pm$0.1 & 53.2$\pm$0.6 & 22.3$\pm$0.0 & 15.9$\pm$0.0 & 34.6$\pm$0.1 \\
    & $M$       & \textbf{48.8$\pm$0.1} & \textbf{35.7$\pm$0.1}  & \textbf{59.6$\pm$0.1} & \textbf{71.7$\pm$0.3} & \textbf{60.6$\pm$0.3}  & \textbf{54.1$\pm$0.6} & \textbf{25.3$\pm$0.1} & \textbf{21.4$\pm$0.1} & \textbf{40.6$\pm$0.3} \\
    & $\neg M$  & 29.9$\pm$0.2 & 20.5$\pm$0.1 & 47.3$\pm$0.2 & 55.4$\pm$0.4 & 41.2$\pm$0.3  & 50.1$\pm$0.4 & 20.3$\pm$0.0 & 12.2$\pm$0.0 & 29.2$\pm$0.2 \\ \hline
    \multirow{3}{*}{HyDE} & ALL                    & 46.7$\pm$0.1 & 32.1$\pm$0.0 & 56.7$\pm$0.1 & 70$\pm$0.1   & 56.7$\pm$0.1 & 54.6$\pm$0.3 & 22.3$\pm$0.0 & 17$\pm$0.0 & 36.6$\pm$0.4 \\
     & $M$ & \textbf{50.2$\pm$0.1} & \textbf{34.5$\pm$0.0} & \textbf{58.8$\pm$0.0} & \textbf{76.5$\pm$0.3} & \textbf{64.8$\pm$0.2} & \textbf{57.7$\pm$0.2} & \textbf{23.6$\pm$0.0} & \textbf{19.5$\pm$0.0} & \textbf{38.1$\pm$0.2} \\
     & $\neg M$       & 29.4$\pm$0.4 & 20$\pm$0.3 & 44.2$\pm$0.2 & 60.5$\pm$0.4 & 44.9$\pm$0.4 & 49.5$\pm$0.3 & 19.4$\pm$0.1 & 11.8$\pm$0.1 & 31$\pm$0.1   \\ \hline
    \end{tabular}}
    \caption{GPT-4o-mini}
    \vspace{2mm}
\end{subtable}
\hfill
\begin{subtable}{0.99\textwidth}
    \centering
    \resizebox{\textwidth}{!}{
    \begin{tabular}{cc|ccc|ccc|ccc}
    \hline
    \multirow{2}{*}{Method} & \multirow{2}{*}{Data} & \multicolumn{3}{c|}{FEVER} & \multicolumn{3}{c|}{SciFact} & \multicolumn{3}{c}{AVeriTeC}  \\ \cline{3-11}
    && Recall@10 & NDCG@10 & F1 & Recall@10 & NDCG@10 & F1 & METEOR & BERTScore & F1\\ \hline
    \multirow{3}{*}{Query2doc} & ALL  & 43.4$\pm$0.1 & 31$\pm$0.1 & 56.9$\pm$0.2  & 64.3$\pm$0.2 & 51.5$\pm$0.1 & 53$\pm$0.5 & 22.5$\pm$0.0 & 16$\pm$0.0 & 34.7$\pm$0.1  \\
    & $M$         & \textbf{47.3$\pm$0.1} & \textbf{34.2$\pm$0.1} & \textbf{59.2$\pm$0.1}  & \textbf{70.9$\pm$0.3} & \textbf{59.7$\pm$0.3} & \textbf{55.3$\pm$0.3} & \textbf{25.7$\pm$0.1} & \textbf{21.5$\pm$0.0} & \textbf{40.4$\pm$0.4}  \\
    & $\neg M$    & 30.5$\pm$0.2 & 20.7$\pm$0.1 & 47.3$\pm$0.2  & 55.8$\pm$0.6 & 40.9$\pm$0.4 & 49.3$\pm$0.2  & 20.4$\pm$0.1 & 12.2$\pm$0.1 & 29.3$\pm$0.2  \\ \hline
    \multirow{3}{*}{HyDE} & ALL  & 46.2$\pm$0.0 & 31.4$\pm$0.0 & 56.7$\pm$0.1  & 71.6$\pm$0.2 & 58.3$\pm$0.1 & 55$\pm$0.3 & 22.8$\pm$0.0 & 17.8$\pm$0.0 & 37.6$\pm$0.3  \\
    & $M$              & \textbf{50$\pm$0.1} & \textbf{34.3$\pm$0.1} & \textbf{58.5$\pm$0.1}  & \textbf{78.9$\pm$0.3} & \textbf{67.1$\pm$0.4} & \textbf{58.7$\pm$0.2} & \textbf{24.2$\pm$0.1} & \textbf{20.4$\pm$0.1} & \textbf{38.9$\pm$0.2} \\
    & $\neg M$         & 32.8$\pm$0.3 & 21.4$\pm$0.2 & 46.6$\pm$0.2  & 63$\pm$0.6 & 48.1$\pm$0.6 & 50.5$\pm$0.3  & 20.5$\pm$0.1 & 13.5$\pm$0.1 & 33.6$\pm$0.3 \\ \hline
    \end{tabular}}
    \caption{Claude-3-haiku}
    \vspace{2mm}
\end{subtable}
\hfill
\begin{subtable}{0.99\textwidth}
    \centering
    \resizebox{\textwidth}{!}{
    \begin{tabular}{cc|ccc|ccc|ccc}
    \hline
    \multirow{2}{*}{Method} & \multirow{2}{*}{Data} & \multicolumn{3}{c|}{FEVER} & \multicolumn{3}{c|}{SciFact} & \multicolumn{3}{c}{AVeriTeC}  \\ \cline{3-11}
    && Recall@10 & NDCG@10 & F1 & Recall@10 & NDCG@10 & F1 & METEOR & BERTScore & F1\\ \hline
    \multirow{3}{*}{Query2doc} & ALL              & 43.4$\pm$0.1 & 31.7$\pm$0.0 & 56.9$\pm$0.1  & 64.7$\pm$0.2 & 52.8$\pm$0.1 & 53.1$\pm$0.6 & 22.3$\pm$0.0 & 16.2$\pm$0.0 & 34.7$\pm$0.2  \\
    & $M$         & \textbf{46.3$\pm$0.0} & \textbf{34.1$\pm$0.0} & \textbf{59.8$\pm$0.1} & \textbf{71.4$\pm$0.3} & \textbf{60.2$\pm$0.3} & \textbf{55.3$\pm$0.3} & \textbf{24.4$\pm$0.0} & \textbf{20.4$\pm$0.1} & \textbf{39.5$\pm$0.3}  \\
    & $\neg M$    & 36.7$\pm$0.1 & 26.1$\pm$0.1 & 49.7$\pm$0.1 & 57.9$\pm$0.4 & 45.1$\pm$ 0.3 & 49.9$\pm$0.3 & 20.7$\pm$0.0 & 12.8$\pm$0.0 & 29.9$\pm$0.3  \\ \hline
    \multirow{3}{*}{HyDE}  & ALL                  & 44.2$\pm$0.1 & 29.9$\pm$0.1 & 56.5$\pm$0.1  & 69.8$\pm$0.1 & 56.6$\pm$0.2 & 54.8$\pm$0.3  & 21.6$\pm$0.0 & 16.3$\pm$0.0 & 37.1$\pm$0.8  \\
    & $M$              & \textbf{47.1$\pm$0.1} & \textbf{31.8$\pm$0.1} & \textbf{59.9$\pm$0.0}  & \textbf{75.5$\pm$0.5} & \textbf{63.8$\pm$0.4} & \textbf{58.3$\pm$0.3} & \textbf{22.6$\pm$0.0} & \textbf{18.6$\pm$0.1} & \textbf{38.5$\pm$0.3}  \\
    & $\neg M$         & 37.5$\pm$0.2 & 25.6$\pm$0.1 & 47.6$\pm$0.2  & 67.7$\pm$0.3 & 53.8$\pm$0.2 & 53.4$\pm$0.2  & 20.1$\pm$0.1 & 12.9$\pm$0.1 & 33.1$\pm$0.4 \\ \hline
    \end{tabular}}
    \caption{Gemini-1.5-flash}
    \vspace{2mm}
\end{subtable}
\hfill
\begin{subtable}{0.99\textwidth}
    \centering
    \resizebox{\textwidth}{!}{
    \begin{tabular}{cc|ccc|ccc|ccc}
    \hline
    \multirow{2}{*}{Method} & \multirow{2}{*}{Data} & \multicolumn{3}{c|}{FEVER} & \multicolumn{3}{c|}{SciFact} & \multicolumn{3}{c}{AVeriTeC}  \\ \cline{3-11}
    && Recall@10 & NDCG@10 & F1 & Recall@10 & NDCG@10 & F1 & METEOR & BERTScore & F1\\ \hline
    \multirow{3}{*}{Query2doc} & ALL              & 43.6$\pm$0.1 & 31.3$\pm$0.1 & 56.8$\pm$0.1  & 63.2$\pm$0.3 & 51.2$\pm$0.1 & 53$\pm$0.5 & 22.3$\pm$0.0 & 15.9$\pm$0.0 & 34.5$\pm$0.1  \\
    & $M$         & \textbf{47.2$\pm$0.1} & \textbf{34.3$\pm$0.2} & \textbf{59.5$\pm$0.0}  & \textbf{71.2$\pm$0.3} & \textbf{60.6$\pm$0.4} & \textbf{55.7$\pm$0.4} & \textbf{25.4$\pm$0.1} & \textbf{21$\pm$0.1} & \textbf{38.5$\pm$0.1}  \\
    & $\neg M$    & 35.8$\pm$0.3 & 24.8$\pm$0.3 & 50.4$\pm$0.2 & 53.8$\pm$0.7 & 40.2$\pm$0.5 & 49.3$\pm$0.2 & 20.4$\pm$0.0 & 12.8$\pm$0.1 & 30.7$\pm$0.2   \\ \hline
    \multirow{3}{*}{HyDE} & ALL                   & 46.4$\pm$0.1 & 31.8$\pm$0.1 & 56.6$\pm$0.1  & 70.1$\pm$0.2 & 57$\pm$0.1 & 54.3$\pm$0.5  & 22.4$\pm$0.0 & 17.4$\pm$0.0 & 36.8$\pm$0.3 \\
    & $M$              & \textbf{50$\pm$0.1} & \textbf{34.3$\pm$0.1} & \textbf{59.1$\pm$0.1}   & \textbf{77.7$\pm$0.6} & \textbf{66.1$\pm$0.4} & \textbf{57.2$\pm$0.3} & \textbf{23.7$\pm$0.1} & \textbf{19.9$\pm$0.1} & \textbf{38.9$\pm$0.2}  \\
    & $\neg M$         & 36.7$\pm$0.3 & 24.9$\pm$0.2 & 48.8$\pm$0.2  & 63$\pm$0.7 & 48.6$\pm$0.5 & 51.4$\pm$0.4 & 21$\pm$0.1 & 14.3$\pm$0.1 & 33.3$\pm$0.3 \\ \hline
    \end{tabular}}
    \caption{Llama-3.1-8b-it}
    \vspace{2mm}
\end{subtable}
\hfill
\begin{subtable}{0.99\textwidth}
    \centering
    \resizebox{\textwidth}{!}{
    \begin{tabular}{cc|ccc|ccc|ccc}
    \hline
    \multirow{2}{*}{Method} & \multirow{2}{*}{Data} & \multicolumn{3}{c|}{FEVER} & \multicolumn{3}{c|}{SciFact} & \multicolumn{3}{c}{AVeriTeC}  \\ \cline{3-11}
    && Recall@10 & NDCG@10 & F1 & Recall@10 & NDCG@10 & F1 & METEOR & BERTScore & F1\\ \hline
    \multirow{3}{*}{Query2doc}  & ALL             & 46.1$\pm$0.2 & 33.7$\pm$0.1 & 57.2$\pm$0.2  & 64.6$\pm$0.2 & 52.5$\pm$0.1 & 53.1$\pm$0.3  & 22.7$\pm$0.1 & 16.3$\pm$0.0 & 34.8$\pm$0.1 \\
    & $M$         & \textbf{49.4$\pm$0.2} & \textbf{36.4$\pm$0.1} & \textbf{59.5$\pm$0.0}  & \textbf{72.5$\pm$0.4} & \textbf{61.4$\pm$0.2} & \textbf{55.4$\pm$0.4} & \textbf{25.4$\pm$0.1} & \textbf{21.1$\pm$0.1} & \textbf{39.6$\pm$0.3}  \\
    & $\neg M$    & 34.2$\pm$0.2 & 23.9$\pm$0.3 & 48.1$\pm$0.2  & 54$\pm$0.5 & 40.5$\pm$0.2 & 48.9$\pm$0.2  & 20.2$\pm$0.0 & 11.9$\pm$0.1 & 28.9$\pm$0.3 \\ \hline
    \multirow{3}{*}{HyDE}   & ALL                 & 49.7$\pm$0.2 & 35$\pm$0.2 & 57$\pm$0.1  & 70.8$\pm$0.2 & 57.8$\pm$0.2 & 55$\pm$0.5  & 22.5$\pm$0.2 & 17.5$\pm$0.1 & 36.7$\pm$0.9 \\
    & $M$              & \textbf{54.4$\pm$0.4} & \textbf{38.6$\pm$0.4} & \textbf{59.3$\pm$0.0}  & \textbf{77.5$\pm$0.5} & \textbf{65.6$\pm$0.4} & \textbf{57.8$\pm$0.3} & \textbf{24.7$\pm$0.2} & \textbf{21.3$\pm$0.3} & \textbf{39.6$\pm$0.3}  \\
    & $\neg M$         & 37.9$\pm$0.2 & 25.9$\pm$0.2 & 50.7$\pm$0.2  & 62.7$\pm$0.6 & 48.2$\pm$0.6 & 50.8$\pm$0.3 & 20.7$\pm$0.1 & 14.2$\pm$0.2 & 32.9$\pm$0.4 \\ \hline
    \end{tabular}}
    \caption{Llama-3.1-70b-it}
    \vspace{2mm}
\end{subtable}
\hfill
\begin{subtable}{0.99\textwidth}
    \centering
    \resizebox{\textwidth}{!}{
    \begin{tabular}{cc|ccc|ccc|ccc}
    \hline
    \multirow{2}{*}{Method} & \multirow{2}{*}{Data} & \multicolumn{3}{c|}{FEVER} & \multicolumn{3}{c|}{SciFact} & \multicolumn{3}{c}{AVeriTeC}  \\ \cline{3-11}
    && Recall@10 & NDCG@10 & F1 & Recall@10 & NDCG@10 & F1 & METEOR & BERTScore & F1\\ \hline
    \multirow{3}{*}{Query2doc}  & ALL             & 43.2$\pm$0.2 & 30.8$\pm$0.2 & 56.8$\pm$0.1  & 63.1$\pm$0.2 & 51$\pm$0.1 & 52.6$\pm$0.4 & 22.5$\pm$0.0 & 16.2$\pm$0.0 & 34.6$\pm$0.1  \\
    & $M$         & \textbf{47.5$\pm$0.2} & \textbf{34.3$\pm$0.2} & \textbf{59.2$\pm$0.0} & \textbf{71.5$\pm$0.4} & \textbf{60.8$\pm$0.3} & \textbf{55.1$\pm$0.2} & \textbf{25.2$\pm$0.1} & \textbf{21.1$\pm$0.1} & \textbf{39.8$\pm$0.3}  \\
    & $\neg M$    & 31.8$\pm$0.4 & 21.5$\pm$0.3 & 48.9$\pm$0.1 & 54.3$\pm$0.5 & 40.8$\pm$0.2 & 49.4$\pm$0.3  & 20.4$\pm$0.0 & 12.2$\pm$0.1 & 29.3$\pm$0.2 \\ \hline
    \multirow{3}{*}{HyDE}  & ALL                  & 46.1$\pm$0.1 & 31.2$\pm$0.0 & 56.5$\pm$0.1  & 69.8$\pm$0.2 & 56.4$\pm$0.2 & 54.8$\pm$0.3 & 22.7$\pm$0.0 & 17.5$\pm$0.0 & 37.3$\pm$0.5  \\
    & $M$              & \textbf{50$\pm$0.2} & \textbf{33.9$\pm$0.1} & \textbf{58.9$\pm$0.1}  & \textbf{76.7$\pm$0.4} & \textbf{64.7$\pm$0.4} & \textbf{57.4$\pm$0.7} & \textbf{23.9$\pm$0.1} & \textbf{20.1$\pm$0.1} & \textbf{38.5$\pm$0.4} \\
    & $\neg M$         & 34.3$\pm$0.2 & 23.3$\pm$0.1 & 48.4$\pm$0.2  & 63$\pm$0.2 & 48.5$\pm$0.2 & 51.5$\pm$0.3 & 21.1$\pm$0.1 & 14.3$\pm$0.1 & 34.4$\pm$0.3 \\ \hline
    \end{tabular}}
    \caption{Mistral-7b-it}
    \vspace{2mm}
\end{subtable}
\hfill
\begin{subtable}{0.99\textwidth}
    \centering
    \resizebox{\textwidth}{!}{
    \begin{tabular}{cc|ccc|ccc|ccc}
    \hline
    \multirow{2}{*}{Method} & \multirow{2}{*}{Data} & \multicolumn{3}{c|}{FEVER} & \multicolumn{3}{c|}{SciFact} & \multicolumn{3}{c}{AVeriTeC}  \\ \cline{3-11}
    && Recall@10 & NDCG@10 & F1 & Recall@10 & NDCG@10 & F1 & METEOR & BERTScore & F1\\ \hline
    \multirow{3}{*}{Query2doc}  & ALL             & 43.4$\pm$0.2 & 30.8$\pm$0.2 & 56.8$\pm$0.2  & 63.6$\pm$0.2 & 51.3$\pm$0.1 & 52.8$\pm$0.4  & 22.6$\pm$0.0 & 16.1$\pm$0.0 & 34.9$\pm$0.1 \\
     & $M$         & \textbf{47.3$\pm$0.2} & \textbf{33.9$\pm$0.2} & \textbf{59$\pm$0.1} & \textbf{71.3$\pm$0.5} & \textbf{60.6$\pm$0.4} & \textbf{54.6$\pm$0.6} & \textbf{25.1$\pm$0.1} & \textbf{20.7$\pm$0.1} & \textbf{39$\pm$0.2}  \\
     & $\neg M$    & 28.7$\pm$0.3 & 19.2$\pm$0.2 & 46.7$\pm$0.2  & 53.8$\pm$0.5 & 39.5$\pm$0.3 & 49.2$\pm$0.4  & 20.2$\pm$0.1 & 11.7$\pm$0.1 & 29.5$\pm$0.2 \\ \hline
    \multirow{3}{*}{HyDE}    & ALL                & 47.6$\pm$0.0 & 32.6$\pm$0.0 & 56.9$\pm$0.2   & 70.2$\pm$0.2 & 56.8$\pm$0.1 & 54.8$\pm$0.5  & 22.8$\pm$0.0 & 17.6$\pm$0.0 & 37.2$\pm$0.5 \\
    & $M$              & \textbf{50.9$\pm$0.1} & \textbf{34.9$\pm$0.0} & \textbf{58.8$\pm$0.0}  & \textbf{77.7$\pm$0.1} & \textbf{65.2$\pm$0.2} & \textbf{56.8$\pm$0.3} & \textbf{24.2$\pm$0.0} & \textbf{20.1$\pm$0.1} & \textbf{38.7$\pm$0.3}  \\
    & $\neg M$         & 33.5$\pm$0.3 & 22.8$\pm$0.2 & 47.2$\pm$0.1  & 61$\pm$0.4 & 46.6$\pm$0.3 & 51.7$\pm$0.4 & 21$\pm$0.1 & 14.3$\pm$0.1 & 33.8$\pm$0.4  \\ \hline
    \end{tabular}}
    \caption{Mixtral-8x7b-it}
    \vspace{2mm}
\end{subtable}
\vspace{-2mm}
\caption{Fact verification performance depending on whether the document generated by query expansion methods contains sentences entailed by gold evidence, with the number of retrieved evidence set to ten ($k=10$). We report performance using different backbone LLMs for query expansion.}
\label{tab:app:nli_top_10}
\end{table*}

\paragraph{Manual annotation}
To support the validity of the NLI-based matching algorithm, we conducted manual annotations on a sampled dataset. Two authors independently reviewed documents generated by Query2doc and HyDE for all 500 samples in the AVeriTeC development set, following the guideline presented in Figure~\ref{fig:app:matched_guideline}. A claim was labeled as \emph{matched} if the LLM-generated document contained all or part of any gold evidence. For the backbone LLMs, we used Llama-3.1-70b-it for Query2doc and Claude-3-haiku for HyDE, as these models achieved the best performance for their respective methods. The two annotators achieved a high-level of inter-annotator agreement, with a Cohen's kappa of 0.837 across 1,000 generations. The estimated proportion of matched claims were 50.8\% and 59.6\%, respectively, closely aligning with those from the NLI-based method, with differences falling within the error margin.  

\begin{table}[t]
\begin{subtable}{\linewidth}
\centering
\resizebox{\linewidth}{!}{
\begin{tabular}{cc|ccc}
\hline
Method & Data & METEOR & BERTScore & F1\\ \hline
\multirow{3}{*}{Query2doc} & ALL   & 20.5  & 13.8  & 34.7     \\
 & $M$            & \textbf{21.6}  & \textbf{15.4}  & \textbf{40.9}     \\
 & $\neg M$       & 19.4  & 12.2  & 27.3    \\\hline
\multirow{3}{*}{HyDE} & ALL  & 19.9  & 14.7  & 38.7     \\
 & $M$            & \textbf{20.7}  & \textbf{16.1}  &  \textbf{40.9}    \\
 & $\neg M$       & 18.8  & 12.5  & 33.6    \\\hline
\end{tabular}}
\caption{$k=5$}
\vspace{4mm}
\end{subtable}

\begin{subtable}{\linewidth}
\centering
\resizebox{\linewidth}{!}{
\begin{tabular}{cc|ccc}
\hline
Method & Data & METEOR & BERTScore & F1\\ \hline
\multirow{3}{*}{Query2doc} & ALL   & 23.6  & 17.3  & 36.7     \\
 & $M$            & \textbf{24.9}  & \textbf{19}  & \textbf{45.2}     \\
 & $\neg M$       & 22.2  & 15.5  & 26.6    \\\hline
\multirow{3}{*}{HyDE} & ALL  & 23.9  & 18.2  & 38.3     \\
 & $M$            & \textbf{24.6}  & \textbf{19.7}  &  \textbf{39.4}    \\
 & $\neg M$       & 22.7  & 16.1  &  34.9   \\\hline
\end{tabular}}
\caption{$k=10$}
\end{subtable}
\caption{Fact verification performance on 500 samples from the AVeriTeC development set using manually annotated NLI labels. For query expansion, we used Llama-3.1-70b-it for Query2doc and Claude-3-haiku for HyDE, as each model achieved the best performance for its respective expansion method. GPT-4o-mini was used for verdict prediction.}
\label{tab:app:annotation}
\end{table}

Table~\ref{tab:app:annotation} presents performance depending on manually annotated matching conditions, showing a consistent trend with the results from the NLI-based method (Table~\ref{tab:nli} and Table~\ref{tab:app:nli_top_5}).

\begin{table*}[htbp]
\centering
\begin{subtable}[t]{\textwidth}
    \centering
    \begin{tabularx}{\textwidth}{|c|X|}
        \hline
        \textbf{Claim} & \begin{minipage}{\linewidth}
        \vspace{2pt}
        U.S. President Donald Trump and his wife, First Lady Melania Trump, have tested positive for COVID-19
        \vspace{2pt}
        \end{minipage} \\ \hline
        \textbf{Gold Evidence} & \makecell[l]{\textbullet~\hlcyan{Donald trump tweeted that he had tested positive for Covid 19} \\~~\hlcyan{on 1st October 2020} \\
        \textbullet~\hlyellow{Donald trump tweeted that he and first lady Melania} \\~~\hlyellow{had tested positive for Covid 19 on 1st October 2020} \\
        \textbullet~\hlpurple{The President's physician, Navy Cmdr. Dr. Sean Conley, wrote that} \\~~\hlpurple{he received confirmation of the positive tests on Thursday evening.}
        }\\ \hline
        \textbf{Generation} & \begin{minipage}{\linewidth}
        \vspace{2pt}
        \hlyellow{On October 1, 2020, it was announced that U.S. President Donald Trump and his wife, First Lady Melania Trump, had tested positive for COVID-19.} This news came after one of the President's closest advisors, Hope Hicks, had also tested positive for the virus earlier that day. The President and the First Lady began experiencing symptoms shortly after Hicks' diagnosis and were subsequently tested. Both Trumps received their positive test results in the early hours of October 2, 2020.
        
        Following their diagnosis, the President and the First Lady immediately began self-isolating in the White House. President Trump's physician, Dr. Sean Conley, issued a statement saying that the President and the First Lady would remain in quarantine for the next 14 days, as per Centers for Disease Control and Prevention (CDC) guidelines.
        
        The news of the President's positive test result sparked widespread concern and reaction, with leaders from around the world offering their well wishes for a speedy recovery. The diagnosis also led to increased scrutiny of the White House's handling of the pandemic, particularly with regards to the safety protocols in place at the time.
        
        Fortunately, both President Trump and the First Lady made a full recovery from the virus. President Trump was hospitalized for a few days at Walter Reed National Military Medical Center, where he received treatment and was later discharged on October 5, 2020.
\vspace{2pt}
\end{minipage}\\ \hline
    \end{tabularx}
    \caption{Matched example}
    \label{tab:app:example1}
\end{subtable}
\end{table*}

\begin{table*}[htbp]\ContinuedFloat
\centering
\begin{subtable}[t]{\textwidth}
    \centering
    \begin{tabularx}{\textwidth}{|c|X|}
        \hline
        \textbf{Claim} & \begin{minipage}{\linewidth}
        \vspace{2pt}
        US Supreme Court Justice, Ruth Bader Ginsburg, fought cancer for decades before she passed away in September 2020. 
        \vspace{2pt}
        \end{minipage} \\ \hline
        \textbf{Gold Evidence} & \makecell[l]{
        \textbullet~\hlcyan{Ruth Bader Ginsburg was a U.S. Supreme Court justice,} \\~~\hlcyan{the second woman to be appointed to the position.} \\
        \textbullet~\hlyellow{On September 18, 2020, Justice Ruth Bader Ginsburg,} \\~~\hlyellow{the second woman to serve on the Supreme Court of the United States,} \\~~\hlyellow{passed away at the age of eighty-seven} \\
        \textbullet~\hlpurple{She died of complications from metastatic pancreatic cancer.} \\
        \textbullet~\hlgreen{colon cancer (first diagnosed in 1999), pancreatic cancer (2009)} \\
        } \\ \hline
        \textbf{Generation} & \begin{minipage}{\linewidth}
        \vspace{2pt}
        Ruth Bader Ginsburg, a trailblazing \hlcyan{U.S. Supreme Court Justice}, demonstrated unwavering resilience in the face of adversity, including her prolonged battle with cancer. Throughout her life, she faced numerous health challenges, including \hlgreen{colorectal cancer in 1999}, \hlpurple{pancreatic cancer} in \hlgreen{2009}, and lung cancer in 2018. Furthermore, in 2019, she underwent radiation therapy for a tumor on her pancreas. Despite these setbacks, Ginsburg continued \hlyellow{to serve on the Supreme Court}, inspiring countless individuals with her strength, determination, and unwavering commitment to her work. Tragically, her courageous fight against cancer came to an end \hlyellow{on September 18, 2020}, when she \hlyellow{passed away at the age of 87}, leaving behind an enduring legacy as a champion of gender equality, social justice, and the rule of law.
        \vspace{2pt}
        \end{minipage} \\ \hline
    \end{tabularx}
    \caption{Matched example}
    \label{tab:app:example2}
\end{subtable}
\end{table*}

\begin{table*}[htbp]\ContinuedFloat
\centering
\begin{subtable}[t]{\textwidth}
    \centering
    \begin{tabularx}{\textwidth}{|c|X|}
        \hline
        \textbf{Claim} & \begin{minipage}{\linewidth}
        \vspace{2pt}
        While serving as Town Supervisor on Grand Island, Nebraska, US Nate McMurray voted to raise taxes on homeowners. 
        \vspace{2pt}
        \end{minipage} \\ \hline
        \textbf{Gold Evidence} & \makecell[l]{
        \textbullet~Grand Island raised its tax levy on November 8th 2018. \\
        \textbullet~A Public Hearing was held on Thursday, November 8, 2018 at 7:00p.m. \\~~for the purpose of hearing anyone who wants to comment on the proposed \\~~2019 Town of Grand Island budget. Supervisor Nate McMurray closed \\~~the Public Hearing.\\
        } \\ \hline
        \textbf{Generation} & \begin{minipage}{\linewidth}
        \vspace{2pt}
    After researching Nate McMurray's political career, there is no evidence that he has ever served as Town Supervisor on Grand Island, Nebraska. Grand Island is a city in central Nebraska, not a town, and it does not have a position of Town Supervisor.

    Nate McMurray is a former town councilman in the town of Grand Island, New York, which is located in Erie County, New York, not Nebraska. However, there are no reports or records indicating that McMurray voted to raise taxes on homeowners while serving on the Grand Island, New York town council.

    In conclusion, the claim that Nate McMurray voted to raise taxes on homeowners while serving as Town Supervisor on Grand Island, Nebraska is false. Nate McMurray has never held a political position in the state of Nebraska, and the town of Grand Island, Nebraska does not exist. This appears to be a fabricated claim with no basis in fact. 
    \vspace{2pt}
    \end{minipage} \\ \hline
    \end{tabularx}
    \caption{Unmatched example}
    \label{tab:app:example3}
\end{subtable}
\end{table*}

\begin{table*}[htbp]\ContinuedFloat
\centering
\begin{subtable}[t]{\textwidth}
    \centering
    \begin{tabularx}{\textwidth}{|c|X|}
        \hline
        \textbf{Claim} & U.S. mortalities this year are not materially different from the previous 5 years \\ \hline
        \textbf{Gold Evidence} & \makecell[l]{
        \textbullet~The cumulative number of deaths in 2020 is far greater than \\~~the cumulative number of deaths of any year from 2015 to 2019, \\~~as shown in the figures in link below. \\
        \textbullet~The Historical data on death rate shows that \\~~there were more deaths recorded in the singular year 2020 \\~~compared to the death rate rate in each year from 2015-2019} \\ \hline
        \textbf{Generation} & \begin{minipage}{\linewidth}
        \vspace{2pt} **U.S. Mortality Rates Remain Steady, Showing No Significant Increase from Previous Years**

According to recent data, mortality rates in the United States for this year are not showing any significant deviations from the past five years. Despite concerns about various health crises and trends, the overall number of deaths in the country has remained relatively stable.

When analyzing the data, it becomes clear that the mortality rates for this year are largely in line with the averages seen in the previous five years. This suggests that, despite fluctuations in specific causes of death or regional variations, the overall trend in mortality rates has not undergone a significant shift.

Several factors could be contributing to this stability, including advances in medical technology, improvements in healthcare access and quality, and public health initiatives aimed at reducing mortality rates. Additionally, the country's demographics, such as population size and age distribution, have not changed dramatically enough to impact mortality rates substantially.

While specific causes of death, such as deaths from accidents, diseases, or other factors, may have seen some variation, the overall mortality rate has not been significantly affected. This steady trend is a positive sign, indicating that the country's healthcare system and public health efforts are effectively managing and mitigating mortality risks.

It is essential to note, however, that mortality rates can fluctuate over time due to various factors, including emerging health crises, changes in population demographics, or shifts in healthcare policies. Therefore, continuous monitoring and analysis of mortality data are crucial to identifying potential trends and areas for improvement.

In conclusion, the current mortality rates in the United States are not materially different from the previous five years, indicating a stable trend. This stability is a testament to the country's healthcare system and public health efforts, but it also highlights the need for continued vigilance and monitoring to address potential future challenges. 
    \vspace{2pt}
    \end{minipage} \\ \hline
    \end{tabularx}
    \caption{Unmatched example}
    \label{tab:app:example4}
\end{subtable}
\caption{Examples of generated documents and gold evidence for the target claims. Colored highlights indicate information in the generated documents that overlaps with the gold evidence.}
\label{tab:app:examples}
\end{table*}

\section{Qualitative Analysis}
\label{sec:app:analysis}

Table~\ref{tab:app:examples} presents examples of LLM-generated documents along with gold evidence. In example (a), the claim concerns Nigeria's history, and the gold evidence specifies the period under military rule, which was reproduced in the generated document. Example (b) pertains to U.S. Supreme Court Justice Ruth Bader Ginsberg, where the gold evidence provides her bibliography and medical history. The LLM-generated text includes this information along with specific years. Notably, it also introduces an additional fact about lung cancer, which is not covered by the gold evidence. Examples (c) and (d) illustrate unmatched cases where the generated text contains factual errors.

\end{document}